\documentclass{article}

\PassOptionsToPackage{square, numbers, compress,sort}{natbib}
\usepackage[final]{neurips_2020}

\usepackage[utf8]{inputenc} %
\usepackage[T1]{fontenc}    %
\usepackage{hyperref}       %
\usepackage{url}            %
\usepackage{amsfonts}       %
\usepackage{nicefrac}       %
\usepackage{graphicx}
\usepackage{subfigure}
\usepackage{dsfont}
\usepackage{xspace}
\usepackage{amsmath}
\usepackage{amssymb}
\usepackage{mathtools}

\usepackage{hyperref}
\usepackage{amsfonts}
\usepackage{color}

\usepackage{float}
\usepackage{scalerel,stackengine}
\usepackage{soul}
\usepackage[flushleft]{threeparttable}
\stackMath
\newcommand\reallywidehat[1]{%
\savestack{\tmpbox}{\stretchto{%
  \scaleto{%
    \scalerel*[\widthof{\ensuremath{#1}}]{\kern-.6pt\bigwedge\kern-.6pt}%
    {\rule[-\textheight/2]{1ex}{\textheight}}
  }{\textheight}%
}{0.5ex}}%
\stackon[1pt]{#1}{\tmpbox}%
}

\newcommand{\mH}{\mathcal{H}}

\newcommand{\mL}{\mathcal{L}}

\definecolor{darkgreen}{rgb}{0.0, 0.2, 0.13}

\definecolor{cornellred}{rgb}{0.7, 0.11, 0.11}

\usepackage[e]{esvect} %
\DeclareRobustCommand{\vec}[1]{#1}%

\let\oldPhi\Phi
\DeclareRobustCommand{\Phi}{{\vec {\oldPhi}}}
\let\oldPsi\Psi
\DeclareRobustCommand{\Psi}{{\vec {\oldPsi}}}
\let\oldlambda\lambda
\DeclareRobustCommand{\lambda}{{\vec {\oldlambda}}}

\DeclareMathOperator{\Tr}{Tr}

\DeclareMathOperator{\odesolver}{ODESolve}
\newcommand{\vvv}[1]{{\mathbf{#1}}}
\newcommand{\position}{\vec q}

\newcommand{\epos}{\vec x}
\newcommand{\velocity}{{\dot{ \vec q}}}

\newcommand{\momentum}{\vec p}
\newcommand{\state}{\vec z}

\newcommand{\q}{{\position}}
\newcommand{\p}{{\momentum}}
\newcommand{\x}{{\epos}}
\newcommand{\z}{{\state}}
\newcommand{\y}{{\vec y}}
\usepackage[algo2e,ruled,vlined]{algorithm2e}
\include{pythonlisting}
\usepackage[toc,page,header]{appendix}
\usepackage{minitoc}

\usepackage{amsthm}

\usepackage{wasysym}
\usepackage{pifont}
\usepackage{bbm}
\usepackage{lipsum}

\usepackage{import}
\usepackage{pdfpages}
\usepackage{enumitem}

\usepackage{empheq}
\usepackage[most]{tcolorbox}
\newtcbox{\mymath}[1][]{%
    nobeforeafter, math upper, tcbox raise base,
    enhanced, colframe=black!30!black,
    colback=white!30, boxrule=1pt,
    #1}
\newtcbox{\mywboxtext}{on line,colback=white,colframe=black,size=fbox,arc=3pt,boxrule=0.8pt}
\newcommand{\mywboxmath}[1]{\mywboxtext{$#1$}}
\usepackage{wrapfig}
\usepackage{xcolor}
\definecolor{dark-blue}{rgb}{0.15,0.15,0.4}
\definecolor{medium-blue}{rgb}{0,0,0.5}
\hypersetup{
  colorlinks, linkcolor={dark-blue},
  citecolor={dark-blue}, urlcolor={medium-blue}
}

\usepackage[font={small}]{caption}
\usepackage{etoolbox}
\makeatletter
\patchcmd{\hyper@makecurrent}{%
    \ifx\Hy@param\Hy@chapterstring
        \let\Hy@param\Hy@chapapp
    \fi
}{%
    \iftoggle{inappendix}{%
        \@checkappendixparam{chapter}%
        \@checkappendixparam{section}%
        \@checkappendixparam{subsection}%
        \@checkappendixparam{subsubsection}%
        \@checkappendixparam{paragraph}%
        \@checkappendixparam{subparagraph}%
    }{}%
}{}{\errmessage{failed to patch}}

\newcommand*{\@checkappendixparam}[1]{%
    \def\@checkappendixparamtmp{#1}%
    \ifx\Hy@param\@checkappendixparamtmp
        \let\Hy@param\Hy@appendixstring
    \fi
}
\makeatletter

\newtoggle{inappendix}
\togglefalse{inappendix}

\apptocmd{\appendix}{\toggletrue{inappendix}}{}{\errmessage{failed to patch}}
\apptocmd{\subappendices}{\toggletrue{inappendix}}{}{\errmessage{failed to patch}}

\title{
Simplifying Hamiltonian and Lagrangian Neural Networks via Explicit Constraints
}

\author{%
  Marc Finzi\thanks{Equal contribution.}\\
  New York University\\
  \And
  Ke Alexander Wang\footnotemark[1]\\
  Cornell University\\
  \And
  Andrew Gordon Wilson\\
  New York University\\
}

\begin{document}
\doparttoc %
\faketableofcontents %

\maketitle

\begin{abstract}
Reasoning about the physical world requires models that are endowed with the right inductive biases to learn the underlying dynamics. Recent works improve generalization for predicting trajectories by learning the Hamiltonian or Lagrangian of a system rather than the differential equations directly. While these methods encode the constraints of the systems using generalized coordinates, we show that embedding the system into Cartesian coordinates and enforcing the constraints explicitly with Lagrange multipliers dramatically simplifies the learning problem. We introduce a series of challenging chaotic and extended-body systems, including systems with $N$-pendulums, spring coupling, magnetic fields, rigid rotors, and gyroscopes, to push the limits of current approaches. Our experiments show that Cartesian coordinates with explicit constraints lead to a 100x improvement in accuracy and data efficiency. 
\end{abstract}

\begin{figure}[h]
\centering
    \includegraphics[width=\textwidth]{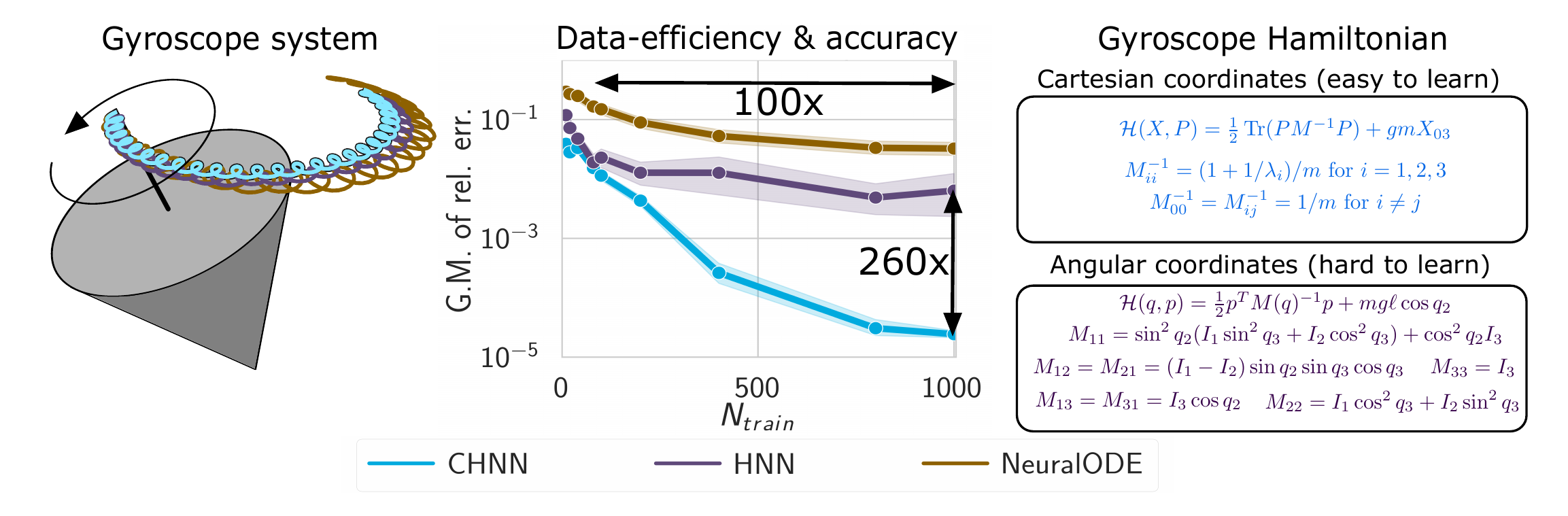}
    \caption{By using Cartesian coordinates with explicit constraints, we simplify the Hamiltonians and Lagrangians that our models learn, resulting in better long term predictions and data-efficiency than Neural ODEs and Hamiltonian Neural Networks (HNNs).
    \textbf{Left:} a spinning gyroscope with the ground truth trajectory and predictions of each model.  Predicted trajectories by our model (CHNN) overlaps almost exactly with the ground truth (black).
    \textbf{Middle:} Geometric mean of the relative error over 100 timesteps as a function of number of training trajectories. On the gyroscope system, our model can be 100 times more data efficient or 260 times more accurate.
    \textbf{Right:} The Hamiltonian expressed in Cartesian coordinates is simpler and easier to learn than when expressed in angular coordinates.}
    \label{fig:frontfig}
\end{figure}

\section{Introduction}
\label{sec:introduction}

\begin{figure}[t]
\centering
    \vspace{-5mm}
    \includegraphics[width=\textwidth]{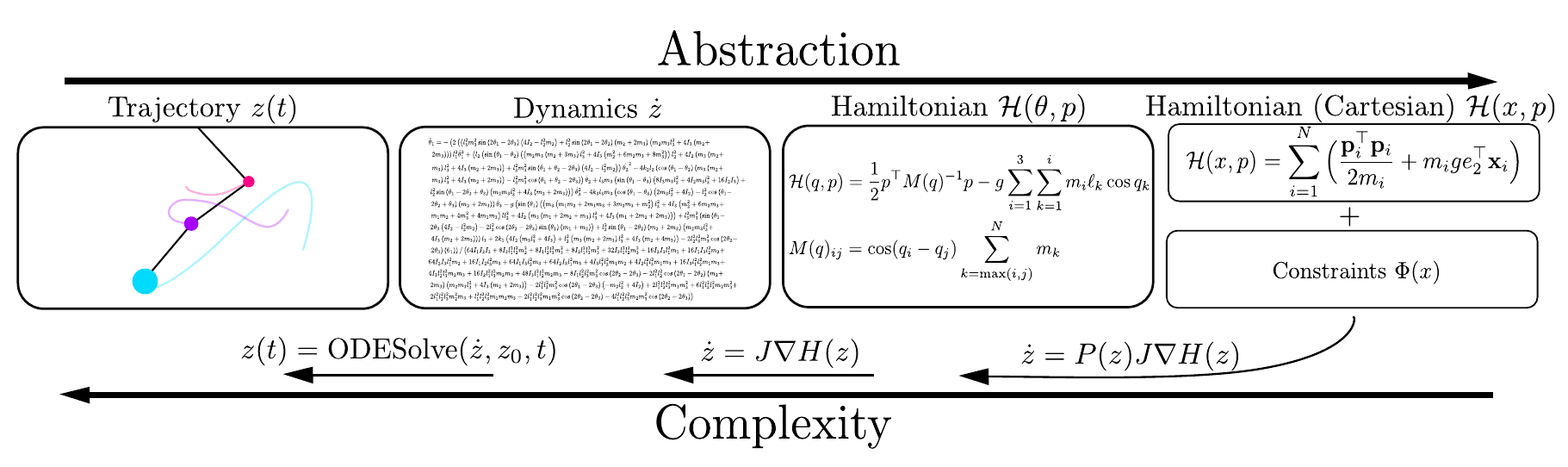}
    \vspace{-5mm}
    \caption{A visualization of how abstracting the physical system reduces the complexity that our model must learn. For systems like the 3-Pendulum, the trajectory is so complex that there is no closed form solution. Although the dynamics $\dot{z}$ do have a closed form, they require a long description. %
    The Hamiltonian \(\mathcal H\) of the system is simpler, and modeling at this higher level of abstraction reduces the burden of learning. Separating out the constraints from the learning problem, the Hamiltonian for CHNNs is even more succinct.
    }
    \label{fig:abstraction}
\end{figure}

Although the behavior of physical systems can be complex, they can be derived from more abstract functions that succinctly summarize the underlying physics. For example, the trajectory of a physical system can be found by solving the system's differential equation for the state as a function of time. For many systems, these differential equations can in turn be derived from even more fundamental functions known as Hamiltonians and Lagrangians. We visualize this hierarchy of abstraction in \autoref{fig:abstraction}.
Recent work has shown that we can model physical systems by learning their Hamiltonians and Lagrangians from data \citep{hnn, delan, zhong2019symplectic}.
However, these models still struggle to learn the behavior of sophisticated constrained systems \citep{hnn,lnn, chen2019symplectic,finzi2020generalizing, garg2019hamiltonian, zhu2020deep}.
This raises the question of whether it is possible to improve model performance by further abstracting out the complexity to make learning easier.

\begin{wrapfigure}{r}{0.5\textwidth}
    \includegraphics[width=0.5\textwidth]{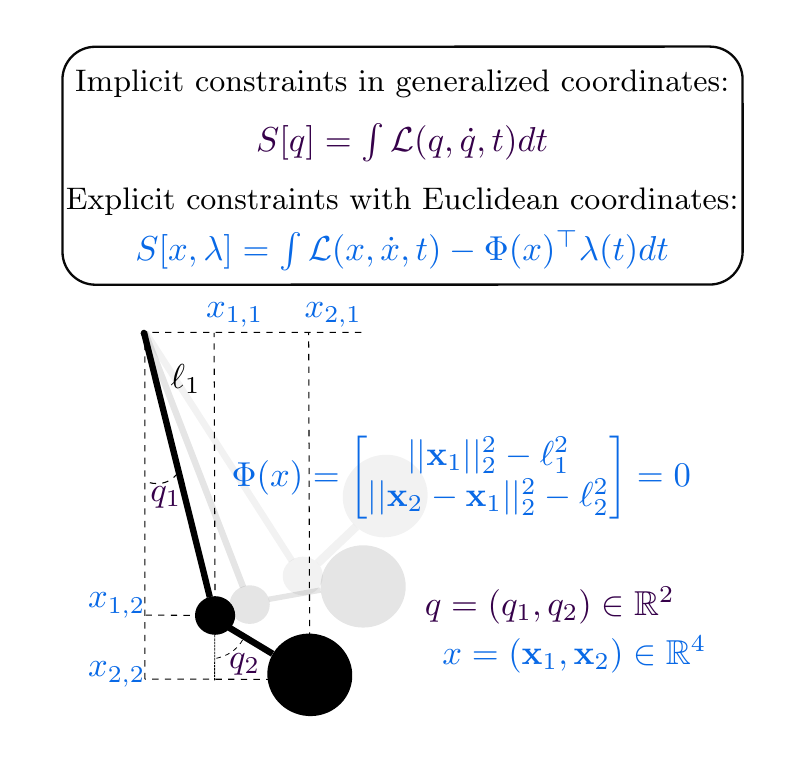}
    \caption{\protect A 2D 2-pendulum expressed in terms of generalized coordinates $\q$ and Cartesian coordinates $\x$ with explicit constraints $\Phi (\x)=\vec{0}$ for the Lagrangian formalism and the constrained Lagrangian formalism. $\mL$ is the Lagrangian, a scalar function that summarizes the entire behavior of the system, entries of $\lambda$ are the Lagrange multipliers, and $S$ is a functional that is minimized by the system's true trajectory.}
    \label{fig:2pendulum-explicit-vs-implicit}
\end{wrapfigure}

Constraints in physical systems are typically enforced by \textit{generalized coordinates}, which are coordinates formed from any set of variables that describe the complete state of the system. For example, the 2D 2-pendulum in \autoref{fig:2pendulum-explicit-vs-implicit} can be described by two angles relative to the vertical axis, labelled as $\q=(q_1, q_2)$, instead of the Cartesian coordinates $\x$ of the masses. By expressing functions in generalized coordinates, we ensure that constraints, like the distances from each pendulum mass to its pivot, are always satisfied implicitly.
However, if we have a mechanism to explicitly enforce constraints, we can instead use Cartesian coordinates, which more naturally describe our three dimensional world.

In this paper, we show that generalized coordinates make the Hamiltonian and the Lagrangian of a system difficult to learn.
Instead, we propose to separate the dual purposes played by generalized coordinates into independent entities: a state represented entirely in Cartesian coordinates $\x$, and a set of constraints $\Phi(\x)$ that are enforced explicitly via Lagrange multipliers $\lambda$. Our approach simplifies the functional form of the Hamiltonian and Lagrangian and allows us to learn complicated behavior more accurately, as shown in \autoref{fig:frontfig}.

In particular, our paper makes the following contributions.
\textbf{(1)} We demonstrate analytically that embedding problems in Cartesian coordinates simplifies the Hamiltonian and the Lagrangian that must be learned, resulting in systems that can be accurately modelled by neural networks with $100$ times less data.
\textbf{(2)} We show how to learn Hamiltonians and Lagrangians in Cartesian coordinates via explicit constraints using networks that we term Constrained Hamiltonian Neural Networks (CHNNs) and Constrained Lagrangian Neural Networks (CLNNs). 
\textbf{(3)} We show how to apply our method to arbitrary rigid extended-body systems by showing how such systems can be embedded purely into Cartesian coordinates.
\textbf{(4)} We introduce a series of complex physical systems, including chaotic and 3D extended-body systems, that challenge current approaches to learning Hamiltonians and Lagrangians. On these systems, our explicitly-constrained CHNNs and CLNNs are $10$ to $100$ times more accurate than HNNs \citep{hnn} and DeLaNs \citep{delan}, which are implicitly-constrained models, and more data-efficient.
Code for our experiments can be found at: \url{https://github.com/mfinzi/constrained-hamiltonian-neural-networks}.

\section{Background on learning dynamical systems}
An ordinary differential equation (ODE) is a system of differential equations which can be described by 
$\dot \z = f(\z,t)$ where $\z(t) \in \mathbb{R}^D$ is the state as a function of time $t$ and $\dot{z}$ is shorthand for $d\z/dt$. $f$ is known as the \textit{dynamics} of the system since it alone specifies how the state changes with respect to time. 
A neural network $f_\theta$ can approximate the dynamics $f$ by learning from trajectory data $\z(t)$ \citep{chen2018neural}. We can make predictions $\hat \z(t)$ by integrating, $\hat{\z}(t)= \odesolver(\z_0,f_\theta,t)$, %
and compute the gradients of the loss function $L(\theta; \z, \hat \z)$ with ordinary backpropagation or the adjoint method \citep{chen2018neural}.

For many physical systems, the differential equations can be derived from one of two scalar functions, a Hamiltonian $\mH$ or a Lagrangian $\mL$, depending on the formalism.
For example, the differential equations of a Hamiltonian system can be written as
\begin{equation}\label{eq:hamiltonian-eom}
    \dot{\z} = J\nabla \mH(\z), \qquad \mathrm{where} \qquad J =
        \begin{bmatrix}
        0 & I_{D/2}\\
        -I_{D/2} & 0
        \end{bmatrix}.
\end{equation}
In this context the state $\z = (\q, \p)$ is a concatenation of the \textit{generalized coordinates} $\q \in \mathbb{R}^{D/2}$ and the \textit{generalized momenta} $\p \in \mathbb{R}^{D/2}$ which parametrize the system's states on a manifold. %
The differential equations of a Lagrangian system can be written in a similar way except that they are only expressed in terms of $\q$ and $\dot \q$. %
Typically, $\dot \q$ and $\p$ are related by $\p = M(\q)\dot \q$ , a generalization of momentum from introductory physics, $\p = m \dot{\x}$, where $M(\q)$ is the \textit{mass matrix}.

Recent approaches predict trajectories by learning $\mH$ or $\mL$ instead of $f$. \citet{hnn} proposed Hamiltonian Neural Networks (HNNs) which parametrize $\mH$ with a neural network. Concurrently, \citet{delan} proposed to learn $\mL$ with Deep Lagrangian Networks (DeLaNs), which was used in robotics applications with additional control inputs. There are two main advantages of this approach: (1) the network only has to learn a scalar function, $\mL$ or $\mH$, whose functional form is simpler than the dynamics $f$, and (2) integrating the differential equations derived from the learned approximations $\mL_\theta$ or $\mH_\theta$ will result in trajectory predictions that better conserve energy since the true dynamics governed by $\mL$ and $\mH$ conserve energy.
Naively, learning $\mH$ requires training trajectories with states $\z=(\q, \p)$. However, in real systems, the states are more commonly available in the form $\z=(\q,\dot{\q})$. This inconvenience can be addressed for most systems since $\p = M(\q)\dot \q$ and we can learn $M$ to convert between $\dot \q$ and $\p$ \citep{zhong2019symplectic}.

\section{Related work}
\label{sec: related}
In addition to the work on learning physical systems above, \citet{chen2019symplectic} showed how symplectic integration and recurrent networks stabilize Hamiltonian learning including on stiff dynamics. \citet{finzi2020generalizing} showed how learned dynamics can be made to conserve invariants such as linear and angular momentum by imposing symmetries on the learned Hamiltonian. \citet{zhong2020dissipative} showed how to extend HNNs to dissapative systems, and \citet{lnn} with LNNs showed how DeLaNs could be generalized outside of mechanical systems such as those in special relativity.

Our method relies on explicit constraints to learn Hamiltonians and Lagrangians in Cartesian coordinates. Constrained Hamiltonian mechanics was developed by \citet{dirac1950generalized} for canonical quantization --- see \citet{date2010lectures} for an introduction. The framework for constrained Lagrangians is often used in physics engines and robotics \citep{murray1994mathematical, featherstone2014rigid} --- see \citet{lavalle2006planning} for an introduction. However, our paper is the first to propose learning Hamiltonians and Lagrangians with explicit constraints. Our approach leads to two orders of magnitude improvement in accuracy and sample efficiency over the state-of-the-art alternatives, especially on chaotic systems and 3D extended-body systems.

\section{Simplifying function approximation with a change of coordinates}\label{sec:simplifying}
Previous works express the position of a system using generalized coordinates $\q$, which has the advantage of automatically satisfying constraints, as explained in \autoref{sec:introduction}.
However, the convenience of using generalized coordinates comes at the cost of making $\mH$ and $\mL$ harder to learn. These complications disappear when we embed the system in the underlying Cartesian coordinates.

We use a simple example to demonstrate how Cartesian coordinates can vastly simplify the functions that our models must learn.
Suppose we have a chain of $N$ pendulums $i=1,...,N$ with point masses $m_i$ in 2D subject to a downward gravitational acceleration $g$. %
Indexing from top to bottom, pendulum $i$ is connected to pendulum $i-1$ by a rigid rod of length $\ell_i$, as shown in \autoref{fig:2pendulum-explicit-vs-implicit}.

In Cartesian coordinates, the Hamiltonian and Lagrangian are simply
\begin{align*}
    \mH(\x, \p) = \sum_{i=1}^N \left[\frac{\vvv{p}_i^\top \vvv{p}_{i}}{2m_i} + m_ige_2^\top \vvv{x}_{i}\right]\quad \text{and} \quad
    \mL(\x, \dot \x) = \sum_{i=1}^N \left[\frac{m_i}{2} \dot{\vvv{x}}_i^\top \dot{\vvv{x}}_i - m_ige_2^\top \vvv{x}_i\right],
\end{align*}
where we used bold to denote the spatial vectors for the position and momentum  $\vvv{x}_i\in \mathbb{R}^2$ and $\vvv{p}_i\in \mathbb{R}^2$ of mass $i$ respectively. Here $\x$ and $\p$ are concatenations of $\vvv{x}_i$ and $\vvv{p}_i$ over $i$.

We can also describe the system with generalized coordinates which implicitly encode the constraints. In this case, %
let $q_i$ be the angle of pendulum $i$ relative to 
the negative $y$ axis and $p_i$ the corresponding generalized momentum.
In these coordinates, the Hamiltonian is
\begin{align}\label{eq:hamiltonian-generalized-coordinates}
    \mH(\q, \p) = \frac 1 2 \p^\top M(\q)^{-1}\p -g\sum_{i=1}^N\sum_{k=1}^i m_i \ell_k\cos{q_k}
\end{align}
where the mass matrix has a complicated form with entries $M(q)_{ij} = \cos(\q_i-\q_j)\ell_i\ell_j\sum_{k=\max(i,j)}^N m_k$ which we derive in \autoref{sec:pendulum}. The Lagrangian is the same as \autoref{eq:hamiltonian-generalized-coordinates} except that the first term is replaced by $\dot\q^\top M(\q)\dot\q/2$ and the second term is negated.

The expression in Cartesian coordinates is \textit{linear} in the state $\x$ and quadratic in $\p$ and $\dot{\x}$ with a \textit{constant} and \textit{diagonal} mass matrix with entries $M_{ii} = m_i$, while the expression in angular coordinates is nonlinear in $\q$ and has off diagonal terms in $M(\q)$ that vary with time as $\q$ varies in time. Moreover, the easiest way to derive the Hamiltonian and Lagrangian in angular coordinates is by first writing it down in Cartesian coordinates and then writing $\x$ in terms of $\q$. This difference in functional form is even more drastic in 3-dimensions where the Cartesian expression is identical, but the Hamiltonian and Lagrangian are substantially more complex. In \autoref{sec:generalized_derivation} we derive additional examples showcasing the complexity difference between coordinates that implicitly enforce constraints and Cartesian coordinates. The constant mass matrix $M$ is in fact a general property of using Cartesian coordinates for these systems as shown in \autoref{sec:embedding3d}.
By simplifying the functional form of $\mH$ and $\mL$, we make it easier for our models to learn.

\section{Learning under explicit constraints}
Although Cartesian coordinates reduce the functional complexity of the Hamiltonian and Lagrangian, they do not encode the constraints of the system. Therefore, we enforce the constraints explicitly for both Hamiltonian dynamics and Lagrangian dynamics using Lagrange multipliers.

\textbf{Hamiltonian mechanics with explicit constraints.}
The dynamics of a Hamiltonian system can be derived by finding the stationary point of the action functional\footnote{Which is in fact exactly the Lagrangian action of the original system, see \autoref{sec:constrained_dynamics} for more details.} %
\begin{align}\label{eqn:hamiltonian-action}
    S[\z] = \int \mL(z(t)) dt = -\int \big[\frac{1}{2}\z(t)^\top J \dot{\z}(t) + \mH(\state) \big] dt,
\end{align}
like in Lagrangian mechanics.
Enforcing the necessary condition of a stationary point $\delta S = 0$ \footnote{$\delta S$ is the variation of the action with respect to $\z$, using the calculus of variations.} yields the differential equation of the system $\dot{\z} = J \nabla \mH$ from \autoref{eq:hamiltonian-eom}, which is shown in \autoref{sec:constrained_dynamics}. We can enforce constraints explicitly by turning this procedure into a constrained optimization problem via Lagrange multipliers.

Suppose we have $C$ holonomic\footnote{Holonomic constraints are equations that give the relationship between position coordinates in the system.} constraints $\{\Phi(\epos)_j = 0\}_{j=1}^C$ collected into a vector $\Phi(\epos) = \vec 0$. We can differentiate the constraints to form an additional $C$ constraints that depend on the momentum $\p$, since $\vec 0=\dot{\Phi}=(D\Phi)\dot{\x} = (D\Phi) \nabla_\p \mH$ where $D\Phi$ is the Jacobian of $\Phi$ with respect to $\x$. If we collect $\Phi$ and $\dot \Phi$, we have $\vec 0 = \Psi(\z) = (\Phi, \dot \Phi) \in \mathbb{R}^{2C}$ as the set of $2C$ constraints that we must enforce when finding a stationary point of $S$.
We can enforce these constraints by augmenting the state $\z$ with a vector of time dependent Lagrange multipliers $\lambda(t) \in \mathbb{R}^{2C}$, giving the augmented action
\begin{equation}
    S[\z,\lambda] = -\int \big[\frac{1}{2}\z^\top J\dot{\z} + \mH(\z) + \Psi(\z)^\top \lambda)\big] dt.
\end{equation}
Enforcing $\delta S = 0$ yields the differential equations that describe the state $\z$ under explicit constraints $\Phi(\x)=\vec 0$:
\begin{equation}\label{eq:constrained_hamiltonian_dynamics}
    \dot{\z} = J\big[\nabla \mH(\z) +  (D\Psi(\z))^\top \lambda \big],
\end{equation}
where $D\Psi$ is the Jacobian of $\Psi$ with respect to $\z$. %
Notice that each row $j$ of $D\Psi$ is the gradient of the constraint $\Psi(z)_j$ and is orthogonal to the constraint surface defined by $\Psi(\z)_j = 0$. 
Left multiplying by $(D\Psi)$ to project the dynamics along these orthogonal directions gives $(D\Psi)\dot \state = d\Psi/dt = \vec 0$ which can then be used to solve for $\lambda$ to obtain
    $\lambda = -\big[(D\Psi) J (D\Psi)^\top \big]^{-1}(D\Psi) J\nabla \mH $.
Defining the projection matrix 
$P := I- J(D\Psi)^\top \big[(D\Psi) J(D\Psi)^\top \big]^{-1}(D\Psi)$, satisfying $P^2=P$,
the constrained dynamics of \autoref{eq:constrained_hamiltonian_dynamics} can be rewritten
as 
\begin{equation}\label{eq:projected_hamiltonian_dynamics}
    \mywboxmath{\dot{\z} = P(\z) J \nabla \mH(\z)}.
\end{equation}
\autoref{eq:projected_hamiltonian_dynamics} can be interpreted as a projection of the original dynamics from \autoref{eq:hamiltonian-eom} onto the constraint surface defined by $\Psi(\x) = \vec 0$ in a manner consistent with the Hamiltonian structure.

\textbf{Lagrangian mechanics with explicit constraints.}
We can perform a similar derivation for a constrained system with the Lagrangian $\mL$.
Given $C$ holonomic constraints $\Phi(\x) = \vec 0$, we show in \autoref{sec:constrained_dynamics} that the constrained system is described by
\begin{equation}\label{eq:constrained_lagrangian_dynamics}
     \mywboxmath{\ddot{\epos} = M^{-1}f - M^{-1}D\Phi^\top \big[D\Phi M^{-1}D\Phi^\top\big]^{-1}D\Phi \big[M^{-1}f + D\dot{\Phi}\dot{\epos} \big]}\,,
\end{equation}
where $D\Phi$ is the Jacobian of $\Phi$ with respect to $\x$, $M = \nabla_{\dot \epos}\nabla_{\dot \epos}\mathcal L$ is the mass matrix, and
$f=f_u+f_c$ is the sum of conservative forces and Coriolis-like forces. \footnote{$f_u(\epos,\dot{\epos}) = \nabla_{\epos} \mathcal L$ and $f_c(\epos,\dot{\epos}) = -(\nabla_{\dot \epos} \nabla_{\epos} \mathcal L)\dot{x}$, but $f_c = 0$ in Cartesian coordinates.}%

\textbf{Learning.} To learn $\mH$ and $\mL$, we parametrize them with a neural network and use \autoref{eq:projected_hamiltonian_dynamics} and \autoref{eq:constrained_lagrangian_dynamics} as the dynamics of the system. This approach assumes that we know the constraints $\Phi(\x)$ and can compute their Jacobian matrices. Since mechanical systems in Cartesian coordinates have separable Hamiltonians and Lagrangian with constant $M$, our method can parametrize $M^{-1}$ with a learned positive semi-definite matrix instead of how it is usually done with a neural network \citep{delan,chen2019symplectic}. \footnote{To enforce positive semi-definiteness of $M^{-1}$, we parametrize the Cholesky decomposition of the matrix $M^{-1}$. In practice, we use a more specialized parametrization of $M$ that is block diagonal but still fully general even when we do not know the ground truth Hamiltonian or Lagrangian, shown in \autoref{eq:mass_matrix}}. For CHNN, we convert from $\dot{x}$ to $p$ and back using the learned mass matrix so that the model can be trained from regular position-velocity data.

\section{Embedding 3D motion in Cartesian coordinates} \label{sec:embedding3d}
Our goal is to learn in Cartesian coordinates, but how can we actually represent our systems in Cartesian coordinates?
Point masses can be embedded in Cartesian coordinates by using a diagonal mass matrix, but it is less obvious how to represent extended rigid bodies like a spinning top. Below we show a general way of embedding rigid-body dynamics in an inertial frame while avoiding all non-Cartesian generalized coordinates such as Euler angles, quaternions, and axis angles which are commonly used in physics simulations \citep{featherstone2014rigid,stewart2000rigid}. %
Additionally by avoiding quaternions, Euler angles, and cross products which are specialized to $\mathbb{R}^3$, we can use the same code for systems in any $\mathbb{R}^d$.

\textbf{Extended bodies in $d$-dimensions.}
In Hamiltonian and Lagrangian mechanics, we may freely use any set of coordinates that describe the system as long as the constraints are either implicitly or explicitly enforced. In fact, at the expense of additional constraints, any non-colinear set of $d$ points $\vvv{x}_1,...,\vvv{x}_d$ of a rigid body in $\mathbb {R}^d$ are fixed in the body frame of the object and completely specify the orientation and center of mass of the object. The rigidity of the object then translates into distance constraints on these points.

Given an extended object with mass density $\rho$ that may be rotating and translating in space, coordinates in the body frame $\vvv{y}$ and coordinates in the inertial frame $\vvv{x}$ are related by $\vvv{x} = R\vvv{y} + \vvv{x}_{cm}$, where $R$ is a rotation matrix and $\vvv{x}_{cm}$ is the center of mass of the object in the inertial frame. As shown in \autoref{sec:3dderivation}, the kinetic energy can be written in terms of $R$ and $\vvv{x}_{cm}$ as 
\begin{equation}\label{eq:kinetic_energy}
    T = m \|\dot{\vvv{x}}_{cm}\|^2/2 +m\Tr(\dot{R}\Sigma\dot{R}^\top)/2,
\end{equation}
where $\Sigma=\mathbb{E}[\vvv{y}\vvv{y}^\top]$ is the covariance matrix of the mass distribution $\rho(\vvv y)$ in the body frame.

Collecting points $\{\vvv{y}_i\}_{i=1}^d$ that are fixed in the body frame,
 we can solve $\vvv{x}_i = R\vvv{y}_i + \vvv{x}_{cm}$ for $R$ to obtain the rotation matrix as a function of $\vvv{x}_{cm}$ and $\{\vvv{x}_i\}_{i=1}^d$. We may conveniently choose these $d$ points to be unit vectors aligned with the principal axes that form the eigenvectors of $\Sigma$ written in the inertial frame.
As we show in \autoref{sec:3dderivation}, when these principal axes and the center of mass are collected into a matrix $X = [\vvv{x}_{cm}, \vvv{x}_1, \ldots, \vvv{x}_d] \in \mathbb{R}^{d\times (d+1)}$, we have $R = X\Delta$ where $\Delta = [-\mathbbm{1},I_{d\times d}]^T$. Plugging in $\dot R = \dot X\Delta$ into \autoref{eq:kinetic_energy} and collecting the $\dot{x}_{cm}$ term gives 
\begin{equation} \label{eq:mass_matrix}
    T = \Tr(\dot{X}M\dot{X}^\top )/2 \qquad \mathrm{where} \qquad M = m\begin{bmatrix}
1+\sum_i\lambda_i & -\lambda^\top  \\
-\lambda & \mathrm{diag}(\lambda)
\end{bmatrix},
\end{equation}
where the $\lambda = \mathrm{diag}(\Sigma)$ are the eigenvalues of $\Sigma$ are collected into a vector $\lambda$. Furthermore,
$M^{-1} = m^{-1}\big(\mathbbm{1}\mathbbm{1}^\top  + \mathrm{diag}([0,\lambda^{-1}_1, \ldots, \lambda^{-1}_d]) \big)$.
Finally, the mass matrix of multiple extended bodies is block diagonal where each block is of the form in \autoref{eq:mass_matrix} corresponding to one of the bodies. Our framework can embed bodies of arbitrary dimension into Cartesian coordinates, yielding the primitives Obj$0$D, Obj$1$D, Obj$2$D, Obj$3$D corresponding to point masses, line masses, planar masses, and 3d objects.

\textbf{Rigidity Constraints.}
To enforce the rigidity of the $d+1$ points that describe one of these extended bodies we use the distance constraints $\Phi(X)_{ij} = \|\vvv{x}_i - \vvv{x}_j\|^2 - \ell^2_{ij} = 0$ on each pair. Technically $\ell_{ij} = 1$ for $i=0$ or $j=0$ and $\sqrt{2}$ otherwise, although these values are irrelevant. Given an Obj$d$D in $d$ ambient dimensions, this translates to $n+1 \choose 2$ internal constraints which are automatically populated for the body state $X\in \mathbb{R}^{d\times (n+1)}$.

\textbf{Joints between extended bodies.}
For robotics applications we need to encode movable joints between two extended bodies. Given two bodies $A$, $B$ with coordinate matrices $X_A$ $X_B$, we use the superscripts $A,B$ on vectors to denote the body frame in which a given vector is expressed. A joint between two extended bodies $A$ and $B$ is specified by point at the pivot that can be written in the body frame of $A$ at location $\vvv{c}^A$ and in the body frame of $B$ at location $\vvv{c}^B$. Converting back to the global inertial frame, these points must be the same. This equality implies the linear constraint
$\Phi(X_A,X_B) =  X_A  \tilde{c}^A-X_B\tilde{c}^B =0$ where $\tilde{c} = \Delta \vvv c+\vvv e_0$. In \autoref{sec:3dderivation}, we show how we incorporate axis restrictions, and how potentials and external forces can be expressed in terms of the embedded state $X$.

\textbf{Simple Links.}
For the links between point masses (Obj$0$D) of pendulums, we use the simpler $\Phi(X_A,X_B) = \|X_A - X_B\|^2 - \ell^2_{AB} = 0$ distance constraints on the $d$ dimensional state vectors $X_A$ and $X_B \in \mathbb{R}^{d\times 1}$. Since $P$ from \autoref{eq:projected_hamiltonian_dynamics} depends only on $D\Phi$, the gradients of the constraints, \textit{we need not know $\ell_{AB}$ in order to enforce these constraints, only the connectivity structure.}

\textbf{Summary.}
To a learn a new system, we must specify a graph that lists the objects (of type Origin, Obj$0$D, Obj$1$D, Obj$2$D, Obj$3$D) and any constraint relationships between them (Link, Joint, Joint+Axis). These constraints are then converted into constraint Jacobians $D\Phi, D\Psi$, written out in \autoref{subsec:jacobians} which define the relevant projection matrices. For each body of type ObjND, we initialize a set of positive learnable parameters $m$, $\{\lambda_i\}_{i=1}^n$ which determine a mass matrix $M$ and $M^{-1}$ using \autoref{eq:mass_matrix} and therefore the kinetic energy $T$. 
Combined with a neural network parametrizing the potential $V(X)$, these form a Hamiltonian $\mH(z) = \mH(X,P) = \Tr(PM^{-1}P^\top )/2+V(X)$ or Lagrangian $\mL(X,\dot{X})=\Tr(\dot{X}M\dot{X}^\top )/2-V(X)$, which could be augmented with additional terms to handle friction or controls as done in \citet{zhong2020dissipative}. Finally,   $\dot{z}=P(z)J\nabla\mH$ (\autoref{eq:projected_hamiltonian_dynamics}) and \autoref{eq:constrained_lagrangian_dynamics} define the constrained Hamiltonian and Lagrangian dynamics that are integrated using a differentiable ODE solver.

\section{Experiments}

\begin{figure}[h!]
    \centering
    \subfigure[]{
	    \centering
	    \includegraphics[width=0.13\textwidth]{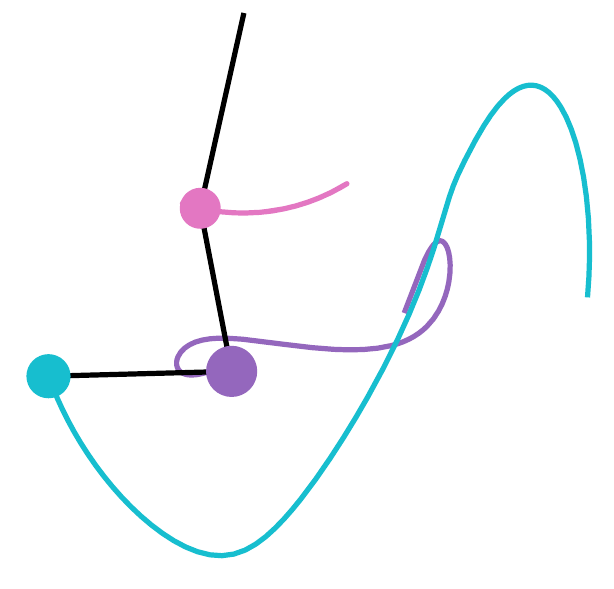}
	    \label{subfig:N-pendulum}
	}
	\subfigure[]{
	    \centering
	    \includegraphics[width=0.20\textwidth]{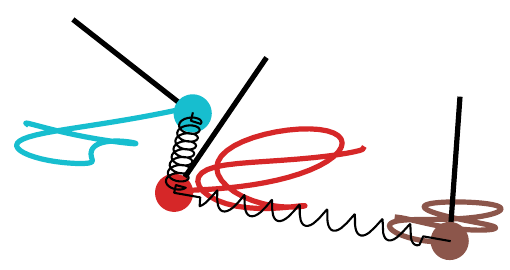}
	    \label{subfig:spring-coupled}
	}
	\subfigure[]{
	    \centering
	    \includegraphics[width=0.14\textwidth]{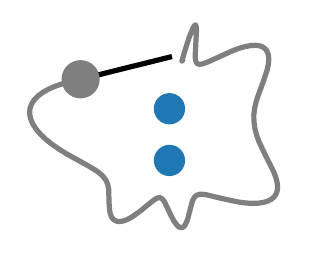}
	    \label{subfig:magnet-pendulum}
	}
	\subfigure[]{
	    \centering
	    \includegraphics[width=0.14\textwidth]{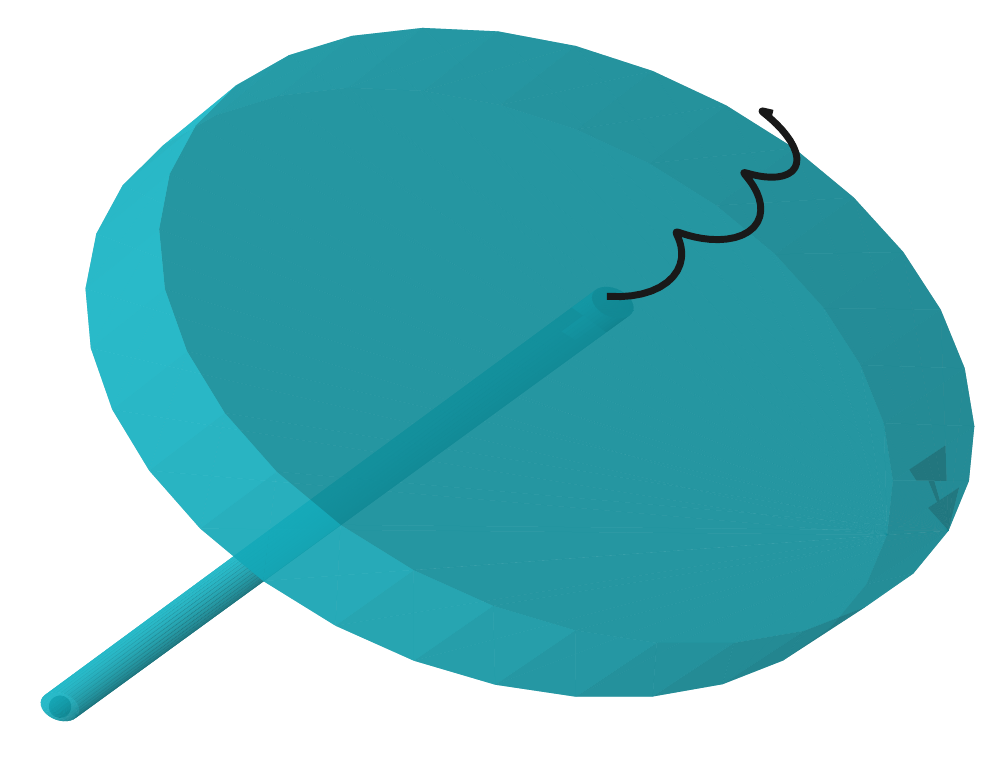}
	    \label{subfig:gyroscope}
	}
    \subfigure[]{
    \centering
	    \includegraphics[width=0.16\textwidth]{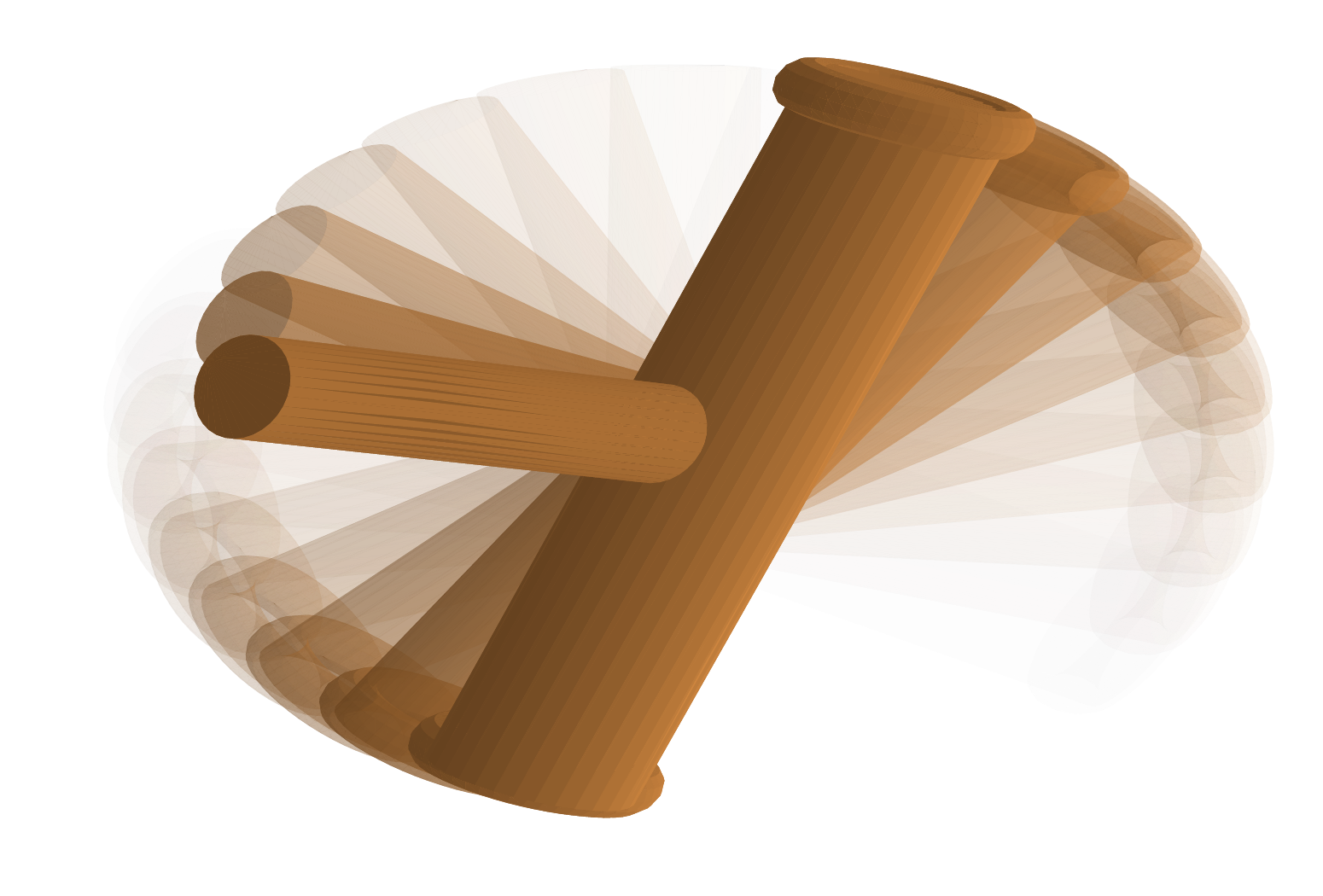}
        \label{subfig:rotor}
	}
	\caption{Systems with complicated dynamics that we simulate. In order from left to right: The N-pendulum, the 3-coupled-pendulum, the magnet pendulum, the gyroscope, and the rigid rotor.}
\end{figure}
\textbf{Datasets and simulated systems.} Previous work has considered relatively simple systems such as the 1 and 2-pendulum \citep{hnn,lnn}, Cartpole, and Acrobot \citep{chen2019symplectic}. We extend the number of links and randomize both masses and joint lengths in the pendulum system to make it more challenging, shown in \autoref{subfig:N-pendulum}. We also add four new benchmarks that test the ability to learn complicated trajectories in 3D. \autoref{subfig:spring-coupled} shows a sequence of 3 pendulums on ball joints that are coupled together with springs, \autoref{subfig:magnet-pendulum} shows a ball joint pendulum with a magnetic tip suspended above two repelling magnets with a complicated potential coming from the magnetic field, inducing chaotic behavior, \autoref{subfig:gyroscope} shows a spinning top which exhibits both precession and nutation, and \autoref{subfig:rotor} shows a free floating rigid rotor with unequal moments of inertia demonstrating the Dzhanibekov effect. \autoref{sec:datasets} describes each system in detail and explains our data generation procedure.

\textbf{Training details.} Following \citep{zhong2019symplectic,sanchez2019hamiltonian,chen2019symplectic} we minimize the error integrated over the trajectories. For each initial condition $(\z_0, t_0)$ in a training minibatch corresponding to a true trajectory $((\z_0, t_0), (\z_1,t_1), \ldots, (\z_n, t_n))$, the model predictions are rolled out with the ODE integrator $(\hat{\z}_1, \hat{\z}_2,..., \hat{z}_n)= \odesolver(\z_0,f_\theta,(t_1,t_2,...,t_n))$
where $f_\theta$ is the learned dynamics.
For each trajectory, we compute the $L_1$ loss averaged over each timestep of the trajectory\footnote{We found that the increased robustness of $L_1$ to outliers was beneficial for systems with complex behavior.} $L(\z, \hat \z) = \frac{1}{n} \sum_{i=1}^n \|\hat{\z}_i-\z_i\|_1$
and compute gradients by differentiating through $\odesolver$ directly. We use $n=4$ timesteps for our training trajectories and average $L(\z, \hat \z)$ over a minibatch of size $200$. To ensure a fair comparison, we first tune all models and then train them for $2000$ epochs which was sufficient for all models to converge. %
For more details on training and settings, see \autoref{sec:architecture}.

\textbf{Evaluating performance.}
We evaluate the relative error between the model predicted trajectory $\hat \state(t)$ and the ground truth trajectory $\state(t)$ over timescales that are much longer than trajectories used at training. Our notion of relative error is $\mathrm{Err}(t) = ||\hat \state(t) - \state (t)||_2 / \left(||\hat\state(t)||_2 + ||\state(t) ||_2\right)$, which can be seen as a bounded version of the usual notion of relative error $||\hat \state(t) -\state(t) ||_2 / ||\state(t)||_2$. $\mathrm{Err}(t)$ measures error independent of the scale of the data and approaches $1$ as predictions become orthogonal to the ground truth or $||\hat \state|| \gg ||\state||$. 
Since the error in forward prediction compounds multiplicatively, we summarize the performance over time by the geometric mean of the relative error over that time interval.
The geometric mean of a continuous function $h$ from $t=0$ to $t=T$ is $\bar{h} = \exp(\int_0^T  \log h(t) dt / T)$, which we compute numerically using the trapezoid rule.
We compare our Constrained Hamiltonian and Lagrangian Neural Networks (CHNNs, CLNNs) against Neural-ODEs \citep{chen2018neural}, Hamiltonian Neural Networks (HNNs) \citep{hnn}, and Deep Lagrangian Networks (DeLaNs) \citep{delan} on the systems described above.
We also evaluate the models' abilities to conserve the energy of each system in \autoref{subsec:energy-conservation} by plotting the relative error of the trajectories' energies.

\textbf{Performance on $N$-pendulum systems.}
The dynamics of the $N$-pendulum system becomes progressively more complex and chaotic as $N$ increases.
\begin{figure}
    \centering
    \includegraphics[width=\textwidth]{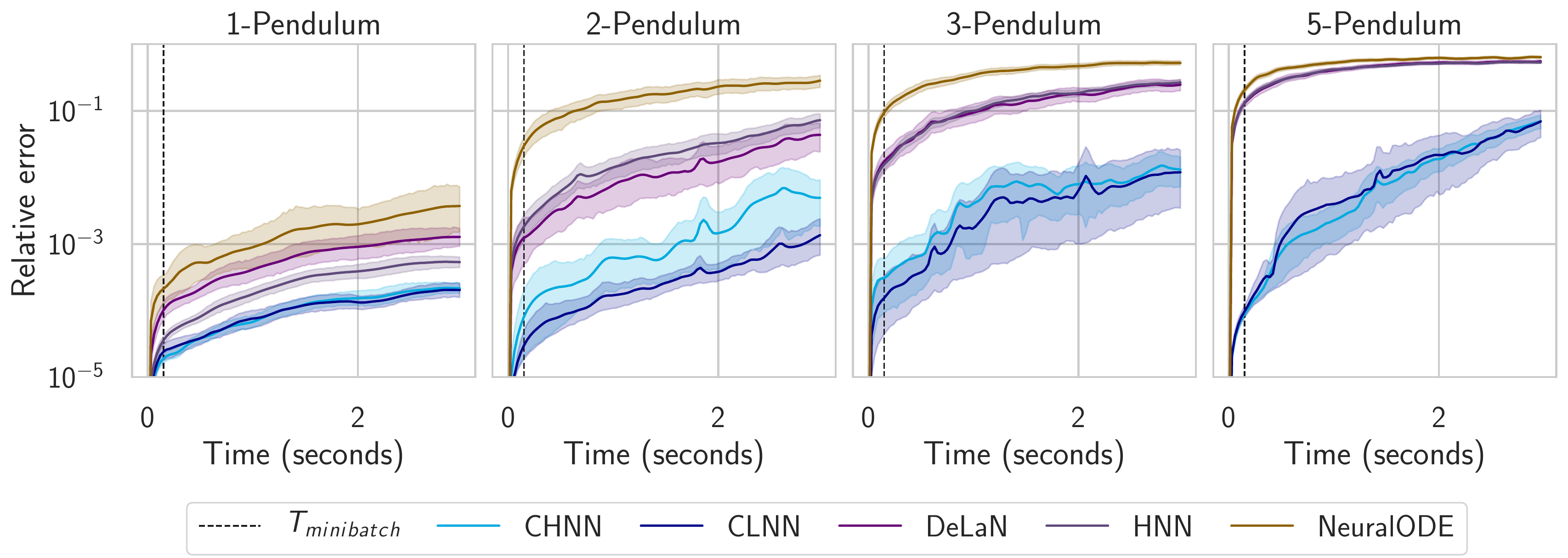}
	\caption{The relative error in the state for rollouts of the baseline NN, HNN, LNN models compared to CHNN and CLNN on the Pendulum Chain tasks. Curves are averaged over $N_{test}=100$ initial conditions and shaded regions are 95\% confidence intervals.
	The vertical axis is log-scaled, meaning that CHNN and CLNN actually have lower variance than the other models. We show this figure in linear scale in \autoref{sec:additional-results}.
	}
	\label{fig:pendulums-rel-err}
\end{figure}
\begin{figure}
    \centering
    \includegraphics[width=\textwidth]{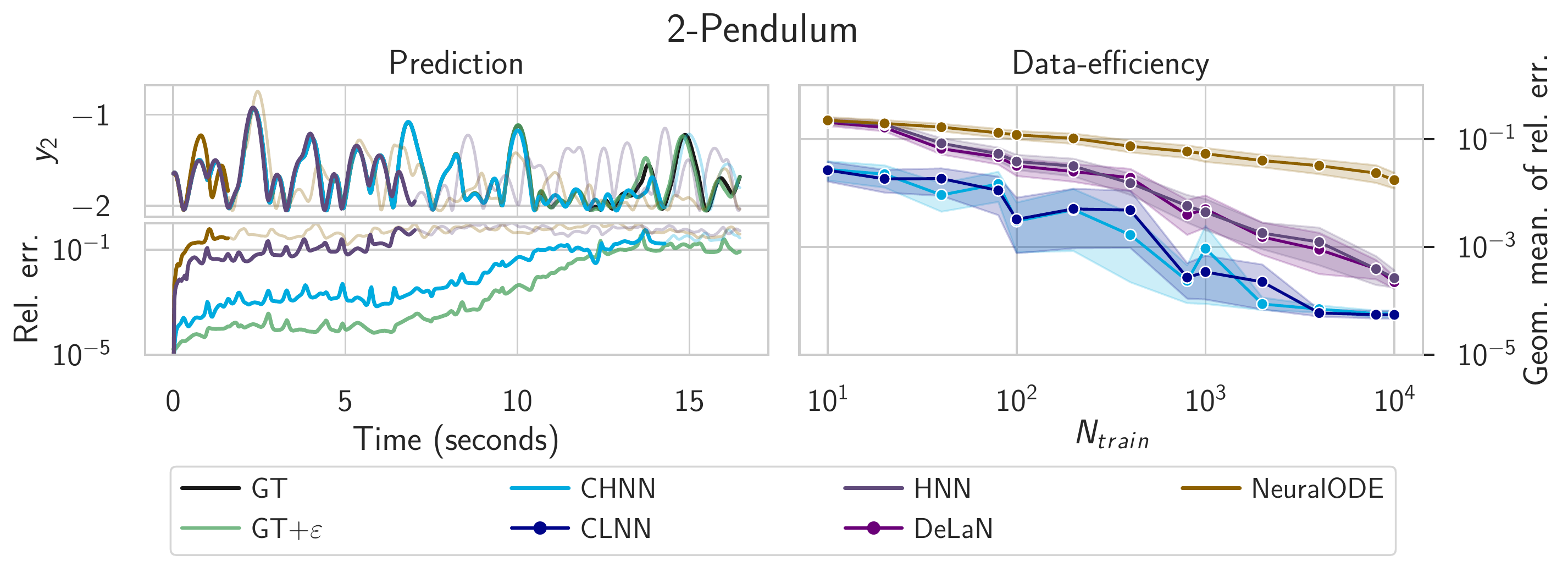}
    \caption{
    \textbf{Left:}  The rollout predictions for the $y$ coordinate of the second bob of a 2-pendulum and relative error over an extended timespan foor CHNN and HNN. Trajectories are faded out several steps after reaching $50\%$ relative error. As the dynamics are chaotic, we also plot a ground truth trajectory that has been perturbed by $\varepsilon=10^{-5}$ showing the natural chaotic growth of error.%
    \textbf{Right:} CHNN and CLNN can achieve the same performance with significantly less data than the baselines. Curves are averaged over $N_{test}=100$ initial conditions and shaded regions are 95\% confidence intervals.}
    \label{fig:data-efficiency}
\end{figure}
For each model, we show its relative error over time averaged over the $N_{test}=100$ initial conditions from the test set in \autoref{fig:pendulums-rel-err} with $N=1,2,3,5$.
Note that each training trajectory within a minibatch for the $N$-pendulum systems is only $T_{minibatch} = 0.12s$ long whereas \autoref{fig:pendulums-rel-err} evaluates the models for $3s$.
All models perform progressively worse as $N$ increases, but 
CHNN and CLNN consistently outperform the competing methods with an increasing gap in the relative error as $N$ increases and the dynamics become increasingly complex.%
We present \autoref{fig:pendulums-rel-err} in linear scale in \autoref{sec:additional-results}, emphasizing that CHNN and CLNN have lower variance than the other methods.

\autoref{fig:data-efficiency} left shows the quality of predictions on the 2-pendulum over a long time horizon of $15s$ with the $y$ coordinate of the second mass for a given initial condition.
As the trajectories for $N$-pendulum are chaotic for $N \geq 2$, small errors amplify exponentially. Even a small perturbation of the initial conditions integrated forward with the ground truth dynamics leads to noticeable errors after $15s$. Notably, our models produce accurate predictions over longer timespans, generalizing well beyond the training trajectory length of $T_{minibatch} = 0.12$ seconds.

\textbf{Data-efficiency.}
As shown in \autoref{sec:simplifying}, the analytic form of the Hamiltonian, and the Lagrangian, are overwhelmingly simpler in Cartesian coordinates. Intuitively, simpler functions are simpler to learn, which suggests that our explicitly-constrained models should be more data-efficient.
\autoref{fig:data-efficiency} right compares the data-efficiency of the models on the chaotic 2-pendulum task. We choose this system because it has been evaluated on in prior work \citep{lnn}, and it is one that previous models can still learn effectively on. %
Our CHNN and CLNN models achieve a lower geometric average error rate using $N_{train}=20$ trajectories than HNN with $N_{train}=200$ trajectories and NeuralODE with $N_{train}=1600$ trajectories.

\begin{figure}
    \centering
    \includegraphics[width=\textwidth]{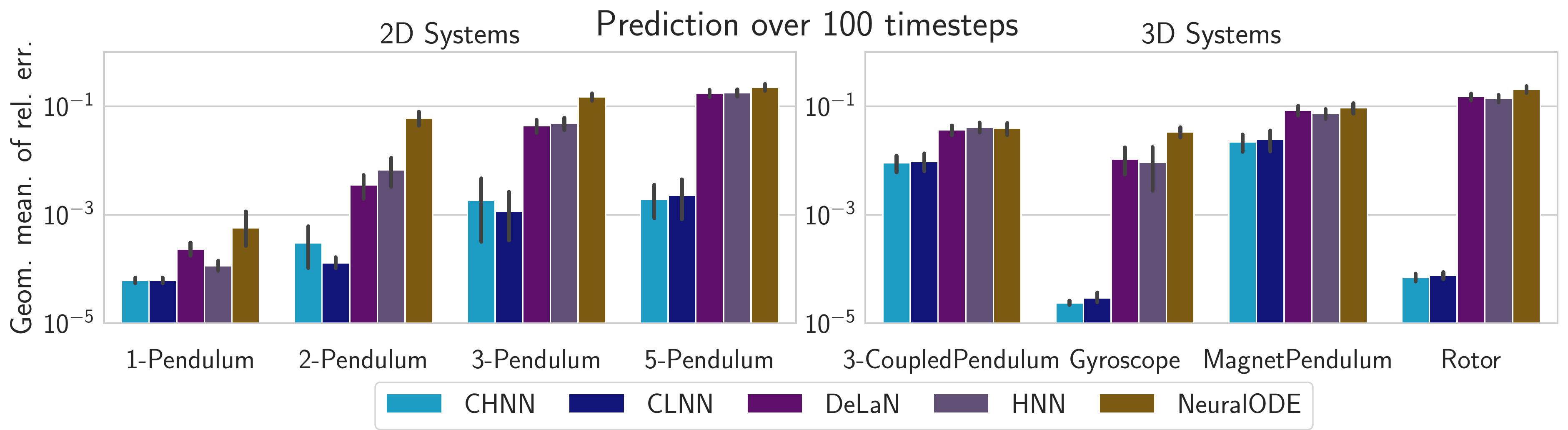}
    \caption{The geometric mean of relative error averaged over $N_{test}=100$ initial conditions with 95\% confidence interval error bars. Models with explicit constraints outperform competing methods on all systems. \textbf{Left}: The 2D $N$-pendulum systems which are chaotic for $N\geq 2$. %
    \textbf{Right}: The 3D systems of which the spring-coupled pendulum and the Magnet-Pendulum are chaotic.%
 	}
    \label{fig:geom-mean}
\end{figure}

\textbf{Performance on 3D systems.}
The pure intrinsic constraint generalized coordinate approach for 3D systems must rely on spherical coordinates and Euler angles, which suffer from coordinate singularities such as gimbal lock. These coordinate singularities lead to singular mass matrices which can destabilize training. We had to apply rotated coordinate systems to avoid singularities when training the baselines models as decribed in \autoref{sec:datasets}. In contrast, CHNN and CLNN naturally circumvent these problems by embedding the system into Cartesian coordinates. As shown in \autoref{fig:geom-mean}, CHNN and CLNN outperform the competing methods both on the tasks where the complexity derives primarily from the coordinate system, the Gyroscope and the Rigid rotor, and on the tasks where the complexity comes from the potential: the spring coupling in 3-CoupledPendulum and the magnetic field in MagnetPendulum.

\section{Conclusion}
We have demonstrated that Cartesian coordinates combined with explicit constraints make the Hamiltonians and Lagrangians of physical systems easier to learn, improving the data-efficiency and trajectory prediction accuracy by two orders of magnitude. We have also shown how to embed arbitrary extended body systems into purely Cartesian coordinates. %
As such, our approach is applicable to rigid body systems where the state is fully observed in a 3D space, such as in robotics. However, Cartesian coordinates are only possible for systems in physical space, which precludes our method from simplifying learning in some Hamiltonian systems like the Lotka-Volterra equations.

There are many exciting directions for future work. Our approach is compatible with recent works that learn dynamical systems with controls \citep{delan,zhong2019symplectic} and in the presence of dissipative forces \citep{zhong2020dissipative}. While we develop the method in a continuous time, there are circumstances involving collision, contacts, and friction where discrete time would be advantageous. Although we used the explicit constraint framework only for Cartesian coordinates in this paper, they can also enforce additional constraints in generalized coordinates, allowing us to pick the best coordinate system for the job. We hope that this approach can inspire handling other kinds of constraints such as gauge constraints in modeling electromagnetism.%
 Finally, although our method requires the constraints to be known, it may be possible to model the constraints with neural networks and propagate gradients through the Jacobian matrices to learn the constraints directly from data.%

\section{Broader Impacts}
Being able to model physical systems accurately has broad applications in robotics, model-based reinforcement learning, and data-driven control systems.
A model that can learn the dynamics of arbitrary systems would greatly reduce the amount of expert-time needed to design safe and accurate controllers in a new environment. %
Although we believe that there are many advantages for using generic neural networks in robotics and control over traditional expert-in the-loop modeling and system identification, neural network models are harder to interpret and can lead to surprising and hard-to-understand failure cases. The adoption of neural network dynamics models in real world control and robotics systems will come with new challenges and may not be suitable for critical systems until we better understand their limitations.%

\textbf{Acknowledgements.} This research is supported by an Amazon Research Award, Facebook Research, Amazon Machine Learning Research Award, NSF I-DISRE 193471, NIH R01 DA048764-01A1, NSF IIS-1910266, and NSF 1922658 NRT-HDR: FUTURE Foundations, Translation, and Responsibility for Data Science.

\bibliographystyle{plainnat}
\bibliography{references}

\cleardoublepage

\appendix
\appendixpage
\section{Overview}
\autoref{sec:additional-results} presents additional experimental results of the paper. %
In \autoref{subsec:energy-conservation} we compare how well the predicted trajectories conserve energy as a function of time. In \autoref{subsec:rel-err-linear-scale} we show the relative error on the $N$-Pendulum systems on a linear scale, which emphasizes that trajectories predicted by CHNNs and CLNNs have lower variance.
In \autoref{subsec:violation} we quantify constraint drift and passively enforced constraints. In \autoref{subsec:effective_dimension} we investigate the complexity of the learning task through the lens of effective dimension \citep{maddox2020rethinking}.

\autoref{sec:derivations} presents the derivation of constrained Hamiltonian and Lagrangian mechanics, and the derivations used to embed 3D motion in Cartesian coordinates.

\autoref{sec:implementation} presents the implementation and training details for our method and the baselines, including documentation of the constraint Jacobians in \autoref{subsec:jacobians}.

\autoref{sec:systems} details the 3D systems that make up the new benchmark datasets, including their Hamiltonians.

Finally, \autoref{sec:generalized_derivation} further demonstrates the complexity of generalized coordinates by deriving the Hamiltonians in generalized coordinates for each of these systems.

\section{Additional results}\label{sec:additional-results}

\subsection{Energy Conservation} \label{subsec:energy-conservation}
\begin{figure}[h]
    \includegraphics[width=\textwidth]{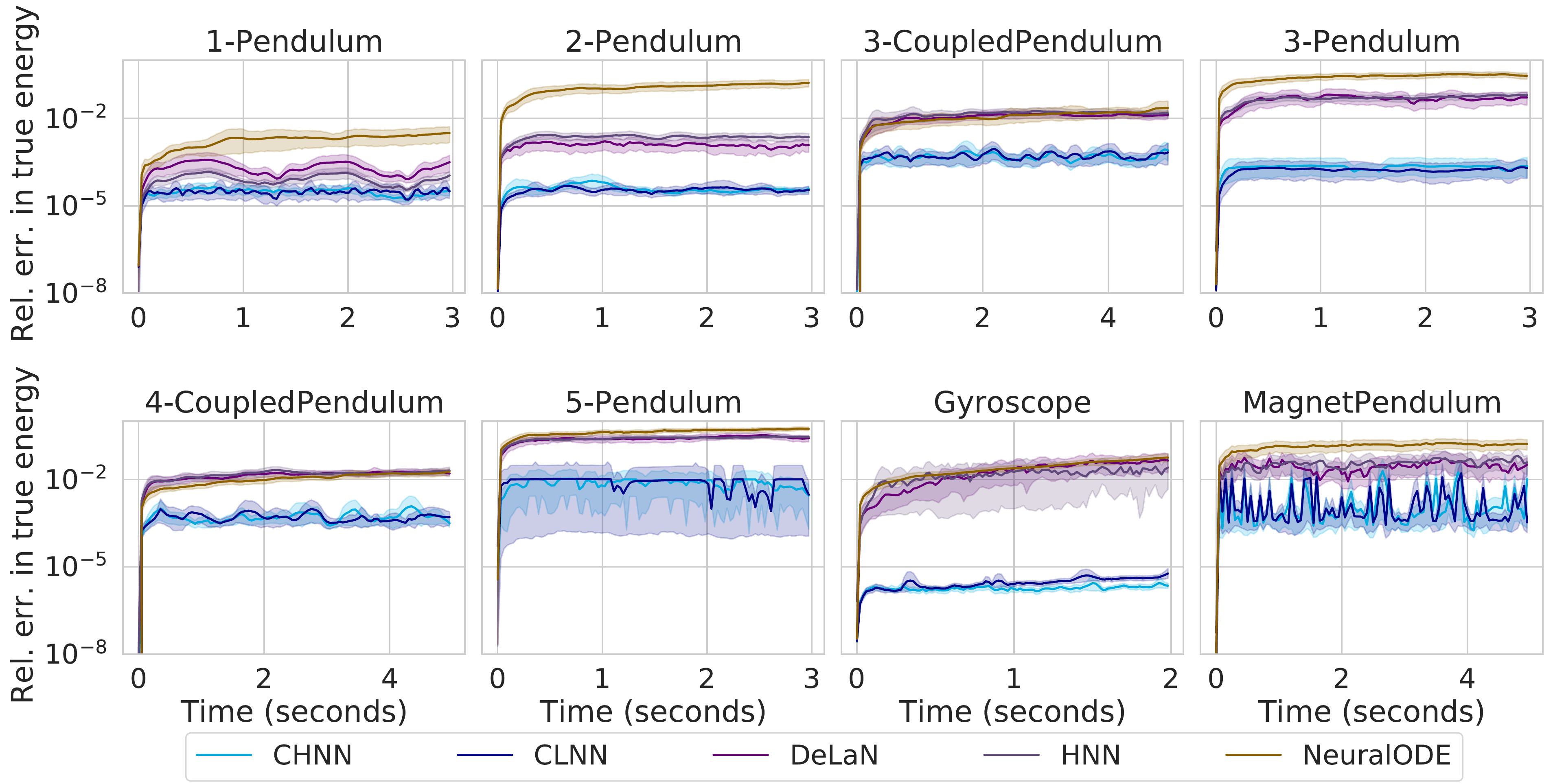}
    \caption{Absolute relative error between the true energy of the predicted trajectories given initial condition $z_0$ and the true energy of the ground truth trajectories starting at $\state_0$. Curves are averaged over $N_{test}=100$ initial conditions and shaded regions are 95\% confidence intervals.}
    \label{fig:energy-conservation}
\end{figure}

When models approximate the true Hamiltonian or the true Lagrangian, they are able to approximately conserve energy. We compare conservation of the true energy of the system of trajectories predicted by each model over time in \autoref{fig:energy-conservation}, showing the relative error $|\mH(\z) - \mH(\hat\z)| / \left(|\mH(\z)| + |\mH(\hat\z)|\right)$.
CHNNs and CLNNs outperform Neural ODEs, HNNs, and DeLaNs on all systems.
All models are able to approximately conserve the true energy of the system on the simple 1-pendulum task but the energy of the trajectories predicted by the baseline models quickly diverge as we transition to the chaotic 3-pendulum and 5-pendulum systems. 

\subsection{Relative error in linear scale}\label{subsec:rel-err-linear-scale}
\begin{figure}
    \centering
    \includegraphics[width=\textwidth]{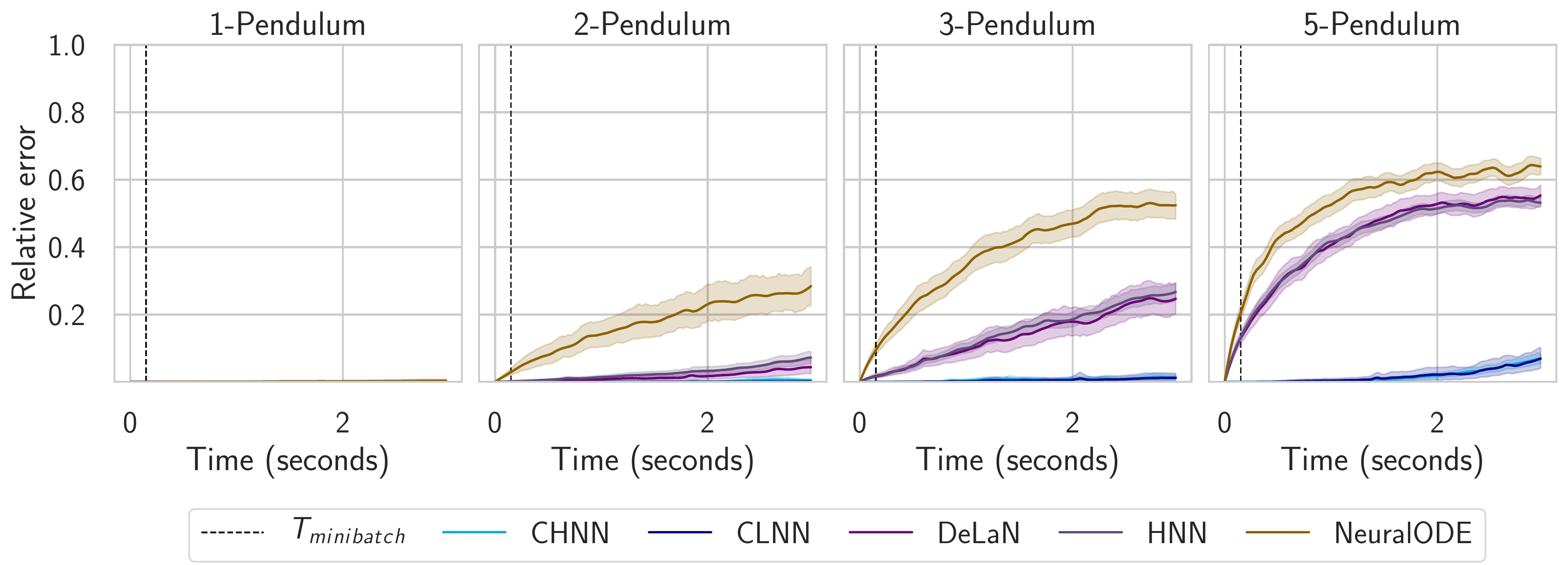}
	\caption{Here we show the linear scale version of \autoref{fig:pendulums-rel-err} which better visually demonstrates the performance difference between our method and the baseline comparisons. The 95\% confidence interval is only perceptively larger for CHNNs and CLNNs in log-scale because they have lower relative error. As shown here, CHNNs and CLNNs in fact have lower variance than the baseline models.}
	\label{fig:pendulums-rel-err-linear}
\end{figure}

We show the relative error of each model on the pendulum systems in linear scale in \autoref{fig:pendulums-rel-err-linear}. We presented the same data in log scale in the main text. By presenting the relative errors in linear scale, it is more visually obvious that CHNNs and CLNNs have lower variance in error than the other methods.

\subsection{Removed constraints and constraint violation.}\label{subsec:violation}
\begin{figure}[h]
    \subfigure[]{
		\includegraphics[width=0.3\textwidth]{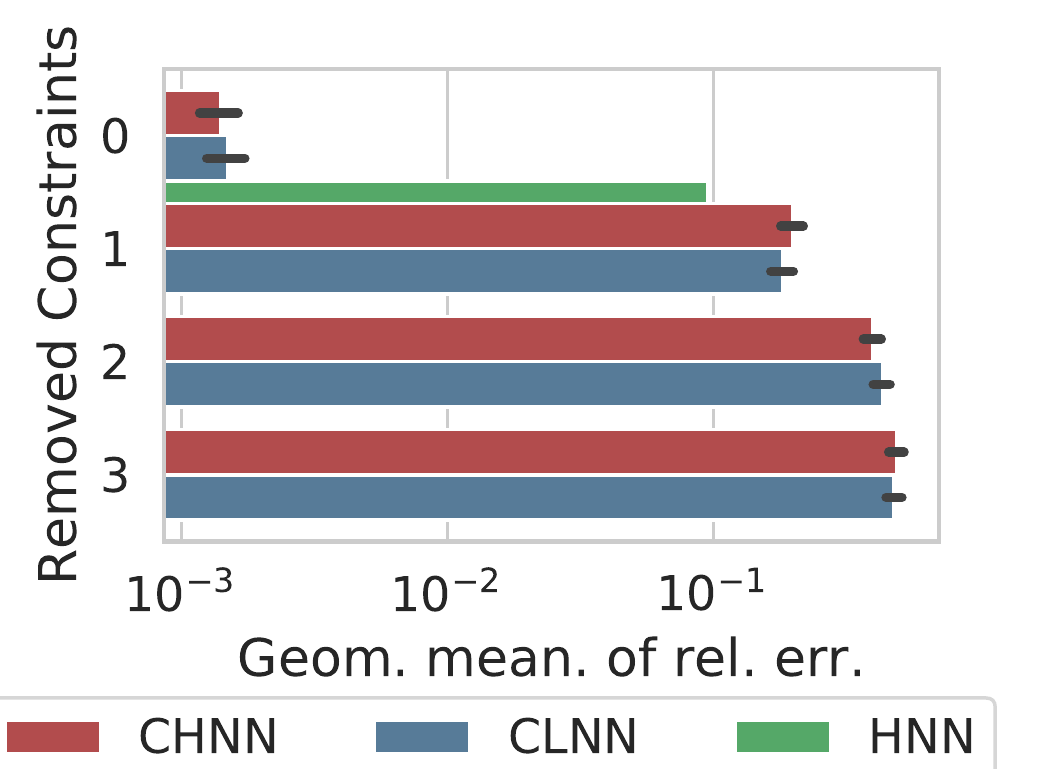}
		\label{subfig:removed_constraints}
    }
        \subfigure[]{
		\includegraphics[width=0.32\textwidth]{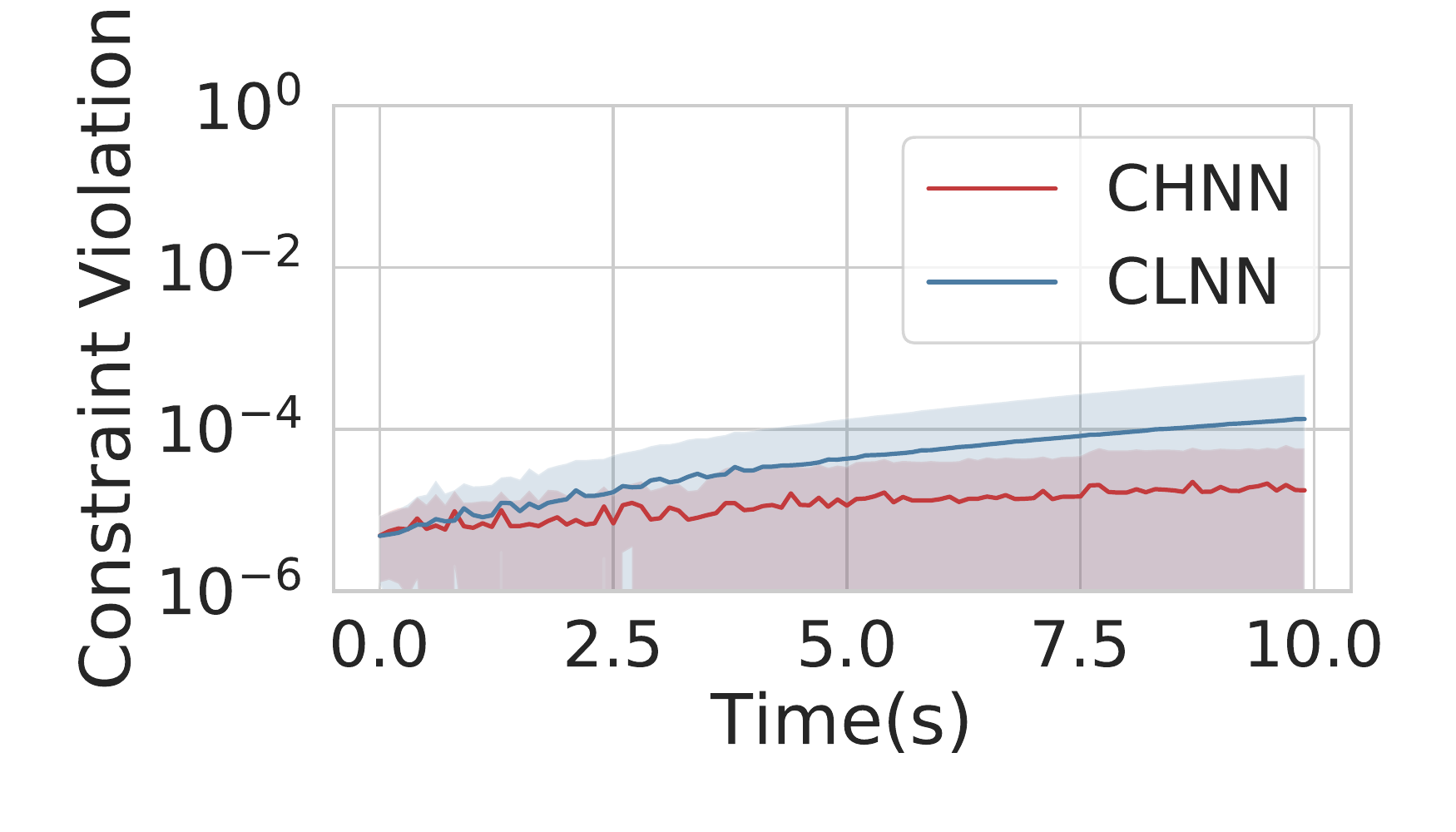}
		\label{subfig:violation}
    }
        \subfigure[]{
		\includegraphics[width=0.32\textwidth]{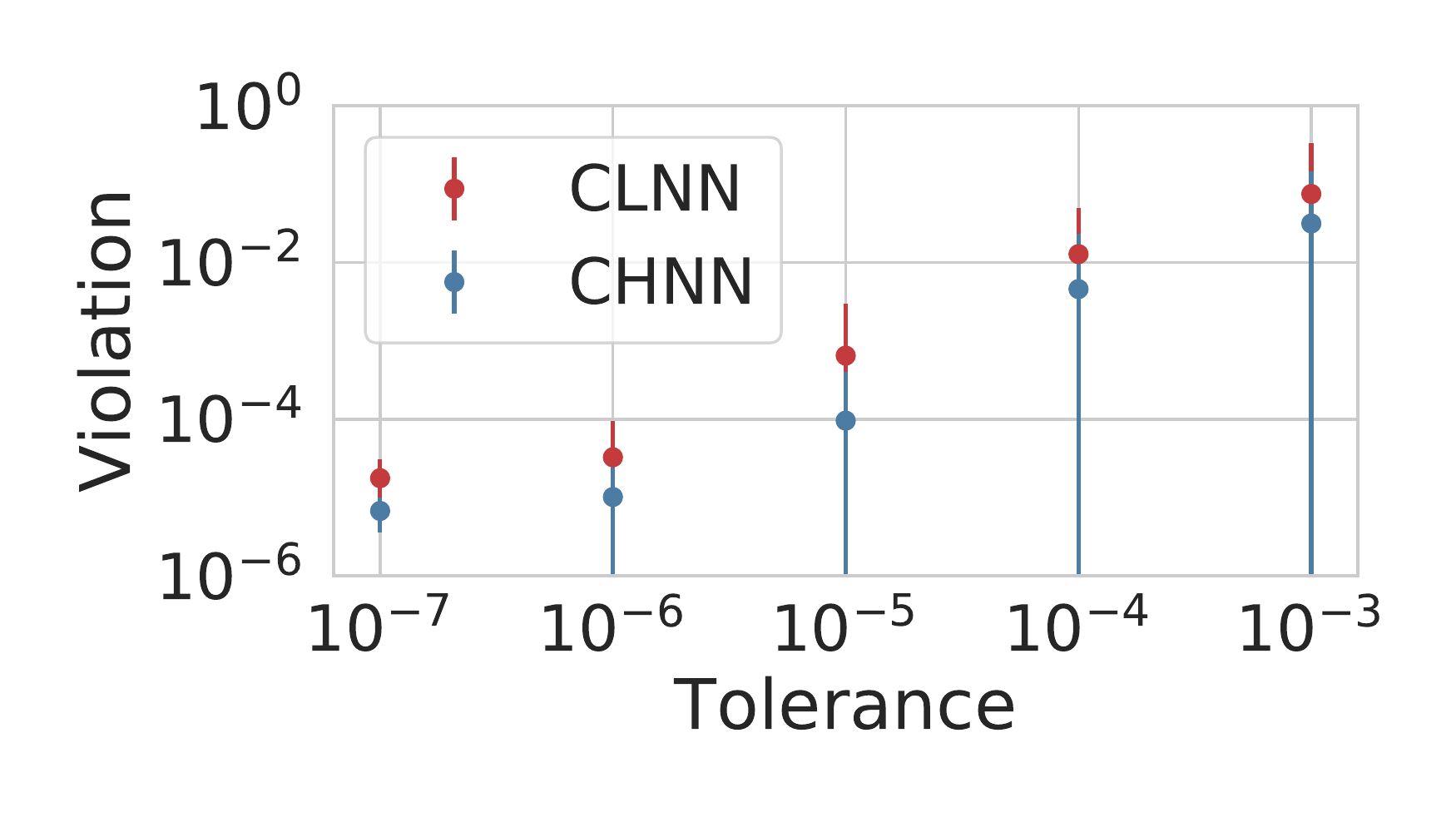}
		\label{subfig:constraint_tolerance}
    }
        
	\caption{\textbf{Left:} We sequentially remove the enforced constraints from the CHNN and CLNN models for the 3-pendulum and evaluate the geometric mean of the relative error against that of a HNN which has all constraints enforced implicitly. Performance degrades rapidly showing that embedding into Cartesian coordinates without explicit constraints is insufficient. \textbf{Middle:} The figure shows how constraint drift leads to a small but increasing constraint violation over time (at $\mathrm{rtol}=10^{-6}$). \textbf{Right:} The geometric mean of the violation over the trajectory is plotted against the relative tolerance of the integrator. By adjusting the tolerance the violation can be controlled. }
    \label{fig:constraints}
\end{figure}

Our method uses Lagrange multipliers to enforce constraints since Cartesian coordinates do not enforce constraints by themselves. In \autoref{subfig:removed_constraints}, we verify that performance rapidly degrades when these constraints are not explicitly enforced for CHNN and CLNNs. In principle the learned Hamiltonian of the model can compensate by approximately enforcing constraints such as with springs, however in practice this does not work well. We compare the performance of our CHNN and CLNN models on the 3-Pendulum (still in Cartesian coordinates) but where the distance constraints are successively removed starting from the bottom. While performance of the base models are >100 times better than HNN, removing even a single constraint results in performance that is slightly worse than an HNN. This experiment shows that both Cartesian coordinates and the explicit enforcement of constraints are necessary for the good performance.

Although the continuous time dynamics of \autoref{eq:projected_hamiltonian_dynamics} and \autoref{eq:constrained_lagrangian_dynamics} exactly preserve the constraints, numerical integration can cause small drifts in the violation of these constraints. The amount of violation can be controlled by the tolerance on the integrator, and we found it to be only a small contribution to the rollout error. 
Symplectic methods for constrained Hamiltonian systems where both phase space area (the symplectic form) and the constraints are exactly preserved by the discrete step integrator have been developed in the literature \citep{leimkuhler1994symplectic} and may prove helpful for very long rollouts.
\autoref{subfig:violation} shows the amount of constraint violation (measured by RMSE on $\Phi$) over time of our models as they predict forward in time. \autoref{subfig:constraint_tolerance} shows the amount of constraint violating as a function of the integrator relative tolerance. As expected, the amount of violation decreases as we make the tolerance more strict.

\subsection{Effective Dimension}\label{subsec:effective_dimension}
We empirically evaluate the complexity of the parameterized Hamiltonian $\mH_\theta$ learned by CHNN and HNN using effective dimensionality \citep{maddox2020rethinking,mackay1992bayesian}. The effective dimensionality of parameterized model $\mH_\theta$ is given by
\begin{align}
    \mathrm{ED}(\mH_\theta) &= \sum_{i=1}^L \frac{\lambda_i}{\lambda_i + z}
\end{align}
where $z$ is a soft cutoff hyperparameter and $\lambda_i$ are eigenvalues of the Hessian of the loss function with respect to the parameters $\theta$. Effective dimensionality characterizes the complexity of a model by the decay of its eigenspectrum. Small eigenvalues with $\lambda_i \ll z$ do not contribute to $\mathrm{ED}(\mH_\theta)$, and eigenvalues above the soft threshold set by $z$ contribute approximately $1$ to $\mathrm{ED}(\mH_\theta)$. 
More complex models have higher effective dimensionality corresponding to slower eigenvalue decay with more eigenvalues greater than the threshold $z$.

\begin{figure}
    \centering
    \includegraphics[width=\textwidth]{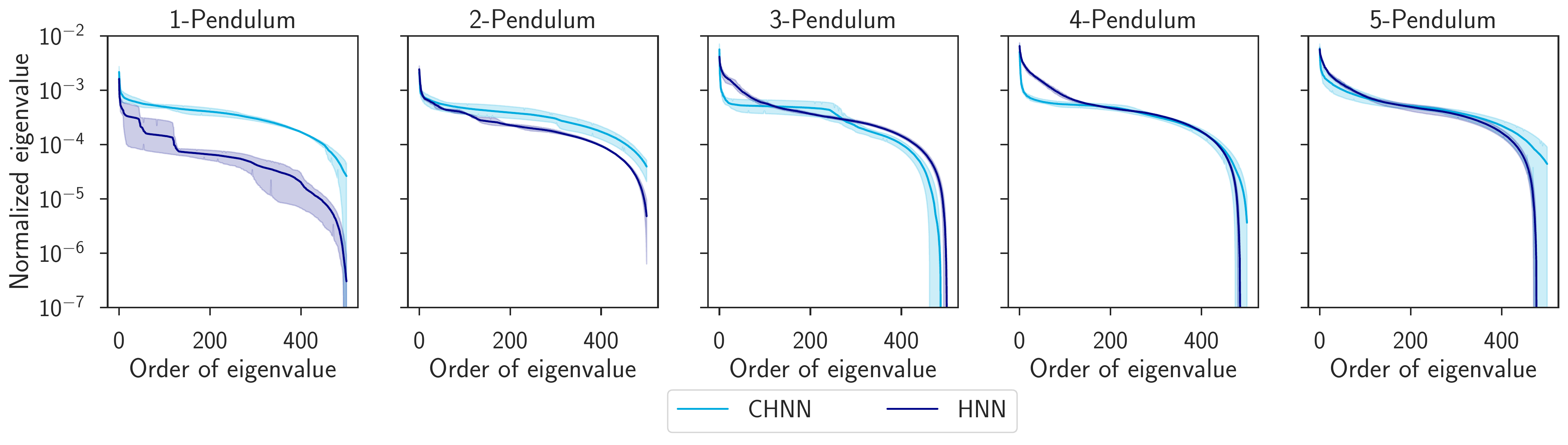}
    \caption{Normalized eigenvalues $\lambda_i / \sum_{i=1}^L\lambda_j$ for the top $N=500$ eigenvalues. Shaded region is 95\% confidence interval averaged over 3 independently trained models. The eigenvalues of CHNN decays more rapidly than those of HNN as the tasks become more difficult, corresponding to a lower effective dimensionality.}
    \label{fig:pendulum-eigenvalues}
\end{figure}

\begin{figure}
    \centering
    \includegraphics[width=0.5\textwidth]{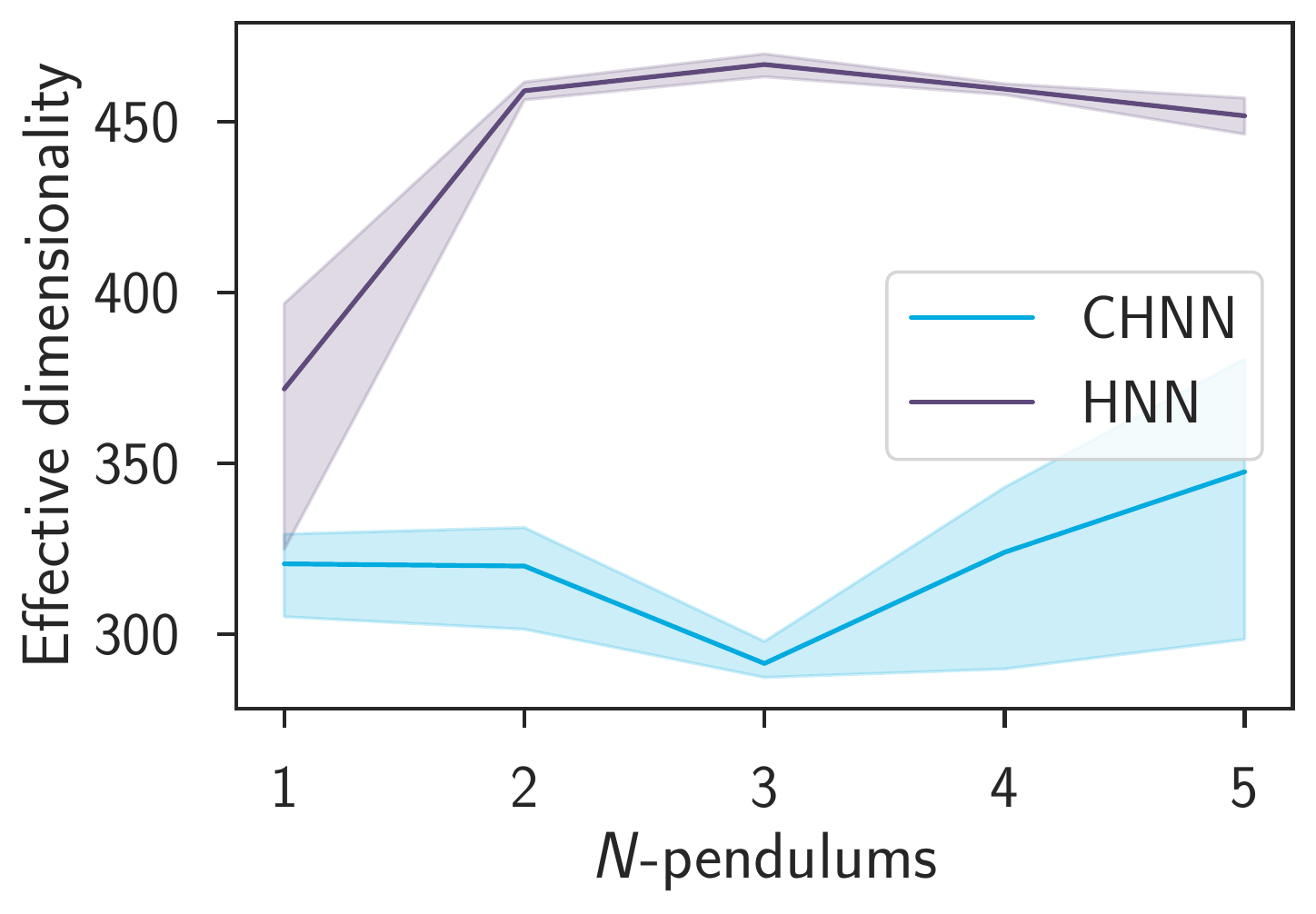}
    \caption{Effective dimensionality of CHNN and HNN on the $N$-Pendulum tasks with $z=10^{-3}$ computed using normalized eigenvalues. Shaded region is 95\% confidence interval averaged over 3 independently trained models.}
    \label{fig:pendulum-ed}
\end{figure}

The loss function differs in scale between CHNN and HNN, with one computing the $L_1$ error between Cartesian coordinates and the other computing the $L_1$ error between angular coordinates, which affects the scale of the eigenvalues. Thus we normalize each eigenvalue by $\sum_{i=1}^L \lambda_i$ before computing the effective dimensionality. In otherwords, we compute effective dimensionality using $\lambda_i \leftarrow \lambda_i / \sum_{j=1}^L \lambda_j$. \autoref{fig:pendulum-eigenvalues} shows the eigenspectra of CHNN and HNN on the $N$-Pendulum systems averaged over 3 runs. Since the models have 133,636 and 269,322 parameters respectively, we only compute the top 500 eigenvalues using the Lanczos algorithm. Based on the elbow in the eigenspectra in \autoref{fig:pendulum-eigenvalues}, we set $z=10^{-3}$ when computing the effective dimensionality shown in \autoref{fig:pendulum-ed}. \autoref{fig:pendulum-ed} shows that the effective dimensionality of HNNs are higher than that of CHNNs which may correspond to the more complex Hamiltonian that HNNs must learn.

\section{Supporting derivations}\label{sec:derivations}
\subsection{Constrained Hamiltonian and Lagrangian Mechanics}\label{sec:constrained_dynamics}
\textbf{Derivation of constrained Hamiltonian mechanics.}
Let $\z = (\epos, \momentum)$ and assume that the Hamiltonian $\mathcal H$ has no explicit dependence on time. The true trajectory $\z$ of a Hamiltonian system is a stationary point of the action functional
\begin{align}\label{eqn:hamiltonian-action}
    S[\z] = -\int \big[\frac{1}{2}\z(t)^\top J \dot{\z}(t) + \mH(\state) \big] dt.
\end{align}
Finding the stationary point by varying the action $\delta S = 0$, %
one recovers the Hamiltonian equations of motion $\dot{\z} = J \nabla \mH$ \citep{date2010lectures}.

\begin{align}
    \delta S &= -\int \big[\frac{1}{2}\delta\z^\top J \dot{\z} +\frac{1}{2}\z^\top J \delta \dot{\z}+ \delta\mH(\state) \big] dt \nonumber\\
    &= -\int \big[\frac{1}{2}\delta\z^\top (J-J^\top) \dot{\z}+ \delta \z^\top \nabla\mH \big] dt \label{eq:by_parts}\\
    &= -\int  \delta\z^\top \big[J \dot{\z} + \nabla\mH \big] dt = 0 \nonumber\\
    J\dot{\z}&= -\nabla \mH\nonumber\\
    \dot{z} &= J \nabla \mH\nonumber
\end{align}
where we use $J=J^\top$ and integrate by parts in \autoref{eq:by_parts}. We use $J^2 = -I$ to obtain the final equation.
In fact, the action is the exact same as the one used to derive the Lagrangian equations of motion. Splitting apart $z=(q,p)$, we have $S = \int \frac{1}{2}(p^\top\dot{q}-q^\top\dot{p}) - H(q,p) dt$ and integrating by parts $S =\int p^\top \dot{q} - H(q,p) dt =  \int \mL(q,\dot{q})dt$.

Now if we have $C$ holonomic constraints given by $\Phi(\epos)_a = 0$ for $a=1,2,...,C$ collected into a vector $\Phi(\epos) = \vec 0$. We can form an additional $C$ constraints that depend on the momentum $\p$, since $\vec 0=\dot{\Phi}=(D\Phi)\dot{\x} = (D\Phi) \nabla_\p \mH$ where $D\Phi$ is the Jacobian of $\Phi$ with respect to $\z$. Collecting both together into a larger column vector $\vec 0 = \Psi(\z) = [\Phi; \dot \Phi] \in \mathbb{R}^{2C}$ gives the set of $2C$ constraints that we must enforce when minimizing \autoref{eqn:hamiltonian-action}.
We can enforce these constraints by augmenting the state $\z$ with a vector of time dependent Lagrange multipliers $\lambda(t) \in \mathbb{R}^{2C}$, yielding the augmented action
\begin{equation*}
    S[\z,\lambda] = -\int \big[\frac{1}{2}\z^\top J\dot{\z} + \mH(\z) +  \Psi(\z)^\top \lambda)\big] dt.
\end{equation*}
Varying the action like before, we get
\begin{equation*}
    \delta S[\z,\lambda] = -\int  \delta\z^\top \big[J \dot{\z} + \nabla\mH  + D\Psi^\top\lambda\big] +\delta \lambda^\top \Psi(z) dt = 0 \\
\end{equation*}
yielding both the constraints and the differential equations that describe the state $\z$ under those constraints:
\begin{equation}\label{eq:constrained_hamiltonian_dynamics_appendix}
    \dot{\z} = J\big[\nabla \mH(\z) +  (D\Psi(\z))^\top \lambda \big] \quad \textrm{and} \quad \Psi(\z)=\vec 0.
\end{equation}
Here $D\Psi$ is the Jacobian of $\Psi$ with respect to $\state$. %

\textbf{Derivation of constrained Lagrangian mechanics.}
We can follow a similar derivation for the constrained Lagrangian formalism. Here, the state is $\state =( \epos, \dot{\epos})$ for a Lagrangian $\mathcal L(\state (t)) = \mL (\position(t), \velocity(t))$. Although the constrained Lagrangian formalism allows for $\velocity$ dependent constraints, known as pfaffian constraints, we will assume a set of $m$ holonomic constraints $\Phi(\position)$ for simplicity. We refer the reader to \citet{lavalle2006planning} on how to extend this framework to allow for pfaffian constraints. 
Like before, since the constraints are not implicit in the coordinate choice, we must enforce them explicitly using Lagrange multipliers. 
Given a set of $m$ holonomic constraints on the position
$\Phi(\epos)_a = 0$ for $a=1,2,...,m$ collected as a column vector $\Phi(\epos) = 0$, one adds a set of time dependent Lagrange multipliers $\lambda_a(t)$ collected as a column vector $\lambda(t)$ to the state with the augmented action:
\begin{align}
    S[\z, \lambda] &= \int \big[\mathcal L(\epos, \dot{\epos}) - \Phi(\epos)^\top \lambda(t)\big] dt.
\end{align}
Enforcing $\delta S/\delta \x=0$  and $\delta S/\delta \lambda = 0$ with vanishing boundary conditions gives $\Phi(\epos) = 0$ and
\begin{equation}\label{eq:constrained_lagrangian_dynamics_appendix}
    \ddot{\epos} = M^{-1}[f -(D\Phi)^\top \lambda]
\end{equation}
where $M = \nabla_{\dot \epos}\nabla_{\dot \epos}\mathcal L$ is the mass matrix and
$f=f_u+f_c$ is the sum of conservative forces $f_u(\epos,\dot{\epos}) = \nabla_{\epos} \mathcal L$ and Coriolis-like forces $f_c(\epos,\dot{\epos}) = -(\nabla_{\dot \epos} \nabla_{\epos} \mathcal L)\dot{x}$ which vanish in Cartesian coordinates.

$\lambda$ can be solved for by noting that $d\Phi/dt = (D\Phi) \dot{\epos} = 0$, and therefore $(D\Phi) \ddot{\epos} = -(D\dot{\Phi}) \dot \epos$ which can be used to substitute for $\ddot{\epos}$. Here $D\Phi$ is the Jacobian of $\Phi$ with respect to $\x$.
Solving for $\lambda$ gives %
\begin{equation}\label{eq:constrained_lagrangian_dynamics_multiplier_appendix}
    \lambda = \big[D\Phi M^{-1}D\Phi^\top\big]^{-1}(D\Phi) \big[M^{-1}f + D\dot{\Phi}\dot{\epos} \big].
\end{equation}
Combining equations \autoref{eq:constrained_lagrangian_dynamics_multiplier_appendix} and \autoref{eq:constrained_lagrangian_dynamics_appendix} then gives the equations of motion in the constrained Lagrangian formalism. %
\subsection{3D systems derivation} \label{sec:3dderivation}
Given an extended object with mass density $\rho$ where the coordinates of points in the body frame $\y$ are related to the coordinates in the inertial frame $\vvv{\x}$ by $\vvv{\x} = R\vvv{\y} + \vvv{\x}_{cm}$, we can split up the kinetic energy into a translational component depending on $\dot{\vvv{\x}}_{cm}$ and a rotational component depending on $\dot{R}$. The mass and center of mass of the object are given by $m=\int d\rho$, $x_{cm} = (1/m)\int x d\rho(x)$.
The kinetic energy of the body is then:
\begin{align}\label{eq:kinetic_energy_appendix}
    T &= (1/2)\int  \dot{\vvv{x}}^\top \dot{\vvv{x}} d\rho(x)
    =(1/2)\int  (\|\dot{\vvv{x}}_{cm}\|^2 + 2\dot{\vvv{x}}_{cm}^\top  \dot{R}\vvv{y}_i + \|\dot{R}\vvv{y}_i\|^2) d\rho\\
    &= (1/2)m\|\dot{\vvv{x}}_{cm}\|^2 + (1/2)\int  \Tr(\vvv{y}\vvv{y}^\top \dot{R}^\top \dot{R})d\rho(y)\\
    &= (1/2)m\|\dot{\vvv{x}}_{cm}\|^2 + (1/2)m\Tr(\dot{R}\Sigma \dot{R}^\top ) = T_{trans}+T_{rot}
\end{align}
where we have defined the matrix of second moments in the body frame $\Sigma = \mathbb{E}[\vvv{\y}\vvv{\y}^\top] = \frac{1}{m}\int \vvv{\y}\vvv{\y}^\top d\rho(y)$, the covariance matrix of the mass distribution, which is a constant. The middle term in \autoref{eq:kinetic_energy_appendix} vanishes since $\mathbb{E}[\vvv{\y}]=(1/m)\int  \vvv{\y} d\rho(y) =0$. This decomposition is exactly mirrors the usual decomposition into rotational energy $T_{rot} = \vvv{\omega}^\top \mathcal{I}\vvv{\omega}$ but is written out differently to avoid angles and specialization to 3D. In 3D, the angular velocity is $\omega = *(\dot{R}R^{-1})$ where $*$ pulls out the components above the diagonal, and the inertia matrix is related to $\Sigma$ by $\mathcal{I} = m(\Tr(\Sigma)I_{3\times 3}-\Sigma)$.

Since the configuration space of a rigid body in $d$ dimensions is $\mathrm{SE}(3)$, we can embed it in a Cartesian space by choosing any $d$ linearly independent points $\{\vvv{y}_i\}_{i=1}^d$ that are fixed in the body frame and for convenience an additional point for the center of mass $\vvv{\y}_{cm}=0$. We may choose these vectors to lie along the principal axis of the object: $\vvv{y}_i = Q_i$ from the eigendecomposition $\Sigma = Q\Lambda Q^\top$. Collecting these points expressed in the body frame into a matrix $X = [\vvv{x}_{cm}, \vvv{x}_1, \ldots, \vvv{x}_d] \in \mathbb{R}^{d\times (d+1)}$, we can express the constraint relating the two frames $R\vvv{y} = \vvv{x} - \vvv{x}_{cm}$ as a linear system:
\begin{align*}
    RQ &= X\begin{bmatrix}
     -\mathbbm{1}^\top  \\ I
     \end{bmatrix} = X\Delta\\
     R &=X\Delta Q^\top\\
     \dot{R} &=\dot{X}\Delta Q^\top
\end{align*}
for the matrix $\Delta = [-\mathbbm{1},I]^\top$.
Combining with \autoref{eq:kinetic_energy_appendix}, the kinetic energy can be rewritten as
\begin{align*}
    T &= m\| \dot{\vvv{x}}_{cm}\|^2/2 + m\Tr(\dot{X} \Delta [Q^\top\Sigma Q] \Delta^\top  \dot{X}^\top )/2 \\
    T &= m\Tr(\dot{X} \big[\vvv{e}_0\vvv{e}_0^\top+ \Delta \Lambda \Delta^\top  \big]\dot{X}^\top )/2,
\end{align*}
where we have made use of $Q^\top\Sigma Q = Q^\top Q \Lambda Q^\top Q = \Lambda$. Or alternatively, we can choose the coordinate system in the body frame so that $Q=I$.
Defining $\Lambda = \mathrm{diag}(\lambda)$ we can collect terms into a single matrix $M$:
\begin{equation} \label{eq:mass_matrix2}
    T(X) = \Tr(\dot{X}M\dot{X}^\top )/2 \qquad \mathrm{where} \qquad M = m\begin{bmatrix}
1+\sum_i\lambda_i & -\lambda^\top  \\
-\lambda & \mathrm{diag}(\lambda)
\end{bmatrix}.
\end{equation}

\textbf{Forces on extended bodies.}
Given a point with components $\vvv{c}\in \mathbb{R}^d$ in the body frame, the vector $x_c$ in the inertial frame has components \begin{equation*}
    \vvv{x}_c = \vvv{x}_{cm}+\sum_i \vvv{c}_i(\vvv{x}_i-\vvv{x}_{cm}) = X[\vvv{e}_0+\Delta \vvv{c}] =X\tilde{c},
\end{equation*} where $\tilde{c} = \vvv{e}_0+\Delta \vvv{c}$. 
Forces $\vvv{f} \in \mathbb{R}^d$ that are applied at that location yield generalized forces on the point collection $F_{k\alpha} = f_k \tilde{c}_\alpha$ for $k=1,2,...,d$ and $\alpha = 0,1...,d$ (in the sense that the unconstrained equations of motion would be $F = \ddot{X}M$). To see this, consider a potential that depends on the location of a certain point $\vvv{x}_c$ of the rigid body $V = V(\vvv{x}_c) =V(X\tilde{c})$. %
Potentials depending on the $\vvv{x}_c$ can be expressed by simply substituting $X\tilde{c}$ in for $x_c$ in the form of the potential.
The (generalized) forces can then be derived via chain rule:
$F_{k\alpha} = \sum_j \frac{\partial V(\vvv{x}_c)}{\partial (\vvv{x}_c)_j} \frac{\partial (\vvv{x}_c)_j}{\partial X_{k\alpha}} =  f_k \tilde{c}_\alpha$. %

\textbf{Rotational axis restrictions.}
Not all joints allow free rotation in all dimensions. Instead it may be that a joint connecting bodies $A,B$ can only rotate about a single axis $\vvv{u}$. In this case, the axis $\vvv{u}$ is fixed in the two frames, and is related by a fixed change of basis when expressed in the body frame of $A$ and the body frame of $B$. This setup gives the constraint
$\Phi(X_A,X_B) = R_A\vvv{u}^A -R_B\vvv{u}^{B} = X_A \Delta \vvv{u}^A - X_B \Delta \vvv{u}^B= 0$, since $R_A = X_A\Delta$, similar to the joint constraint but without the extra $\vvv e_0$.

\section{Implementation details} \label{sec:implementation}
\subsection{Dataset generation}
We generate synthetically datasets using by plugging in known Hamiltonians from a variety of test systems to the framework described above.
For each experiment and each system in \autoref{fig:pendulums-rel-err}, \autoref{fig:geom-mean}, \autoref{fig:energy-conservation}, and \autoref{fig:pendulums-rel-err-linear}, we create a training set by sampling $N_{train}=800$ initial conditions and evaluate the dynamics at 100 timesteps separated by $\Delta t$ specific to the system.
We integrate the dynamics using an adaptive ODE solver, Runge-Kutta4(5), implemented by \citet{chen2018neural} with a relative tolerance of $10^{-7}$ and an absolute tolerance of $10^{-9}$.
We divide each trajectory into 20 non-overlapping chunks each with 5 timesteps. Finally, we choose a chunk at random from each trajectory, resulting in an aggregate training set of $N_{train}=800$ trajectories each with 4 timesteps.
At training time, we create a minibatch by sampling $m=200$ of these shortened trajectories. 
We also create a separate set of $N_{test}=100$ trajectories following the above procedure except that each test trajectory contains the full 100 timesteps without chunking.

For the data-efficiency experiments in \autoref{fig:frontfig} (Middle) and \autoref{fig:data-efficiency}, we generate $N=10000$ training trajectories each with 4 timesteps following the same procedure as the previous paragraph. We then choose the first $N_{train}$ trajectories among these $N$ trajectories as we vary $N_{train}$ on the x-axis. This ensure that the the sequence of training trajectories for each $N_{train}$ is a sequence of monotone increasing sets as $N_{train}$ increases.

\subsection{Architecture}\label{sec:architecture}
For the baseline models which are trained on angular data, care must be taken to avoid discontinuities in the training data. In particular, we unwrap $(\pi-\epsilon \to -\pi +\epsilon)$ to $(-\infty, \infty)$ for the integration and embed each angle $\theta$ into $\sin \theta, \cos \theta$ before passing into the network as in \citet{zhong2019symplectic} and \citet{lnn}, to improve generalization. For all neural networks, we use 3 hidden layers each with 256 hidden units and $\tanh$ activations. We use this architecture to parametrize the potential energy $V$ for all models.
We also use this architecture to parametrize the dynamics for the baseline Neural ODE.

The kinetic energy term $\p^\top M(\q)^{-1}\p/2 = \dot{\q}^\top M(\q)\dot{\q}/2$ is handled differently between our explicitly constrained models and the baselines.
For the baseline HNNs and DeLaNs, we use the above neural network to parameterize the lower triangular matrix $L(\q)$ from the Cholesky decomposition $L(\q)L(\q)^\top = M^{-1}(\q)$,
as done in \citet{delan} and \citet{zhong2019symplectic} since the mass matrix in general is a function of $\q$. 
For CHNNs and CLNNs we parameterize $M$ and $M^{-1}$ using learned parameters $m, \{\lambda_i\}_{i=1}^d$ as described in \autoref{eq:mass_matrix} since $M$ and $M^{-1}$ are constant in Cartesian coordinates.

\subsection{Model selection and training details}
We tune all models on the 3-Pendulum system, which is of intermediate difficulty, using the integrated trajectory loss evaluated on a separate validation set of 100 trajectories. We find that using AdamW \citep{loshchilov2017decoupled} with a learning rate of $3\times 10^{-3}$ and weight decay of $10^{-4}$ along with learning rate cosine annealing \citep{loshchilov2016sgdr} without restarts generally works the best for all models across systems. 
To ensure convergence of all models, we train all models for 2000 epochs even though CHNN and CLNN usually converge within a few hundred epochs.
With the exception of the magnetic pendulum and the rigid rotor, which require a lower learning rate and fewer epochs, we use these settings for all experiments. Despite our best efforts to circumvent this issue, HNN and DeLaN encounter gimbal lock for 3D systems due to their choice of angular coordinates, which causes the loss function to explode if trained for too long. Thus we could only train HNN and DeLaN for 200 epochs on the rigid rotor, which was empirically sufficient to flatten the training loss. We circumvent the coordinate singularity for these baseline models by rotating the coordinate system by $\pi/2$ so that the straight up/straight down configurations do not correspond to a coordinate singularity.

\subsection{Constraint Jacobians}\label{subsec:jacobians}
\textbf{Distance constraints}
For each of the position constraints $\Phi_{ij} = \|\mathbf{x}_i-\mathbf{x}_j\|^2$, there is a conjugate constraint on the velocity:
\begin{equation}
    \dot{\phi}_{ij} = 2(\mathbf{x}_i - \mathbf{x}_j)\cdot (\dot{\mathbf{x}}_i - \dot{\mathbf{x}}_j)  = 0
\end{equation}
Collecting the constraints: $\Phi = [\phi,\dot{\phi}]$ and taking derivatives with respect to $x_{k\ell}$ and $\dot{x}_{k\ell}$, the Jacobian matrix $D_{x,\dot{x}}\Phi \in \mathbb{R}^{2nd \times 2E}$ takes the simple form:
\begin{equation*}
    (D\Phi)_{(k\ell) (nm)} = \begin{bmatrix} \frac{\partial \phi_{nm}}{\partial x_{kl}} & \frac{\partial \dot{\phi}_{nm}}{\partial x_{kl}} \\\frac{\partial \phi_{nm}}{\partial \dot{x}_{kl}} & \frac{\partial \dot{\phi}_{nm}}{\partial \dot{x}_{kl}} \\\end{bmatrix} = \begin{bmatrix}2(x_{kn}-x_{km})(\delta_{n\ell}-\delta_{m\ell}) & 2(\dot{x}_{kn}-\dot{x}_{km})(\delta_{n\ell}-\delta_{m\ell})\\0  & 2(x_{kn}-x_{km})(\delta_{n\ell}-\delta_{m\ell})\end{bmatrix}
\end{equation*}
In $\dot{x}$ is related to $p$ by $\dot{x}_{k\ell} = \frac{\partial \mH}{\partial p_{k\ell}}$ which in general could be a nonlinear function of $x,p$; however, in mechanics $\dot{x}_{k\ell} = \sum_n p_{kn}M^{-1}_{\ell n}$ or in matrix form $\dot{X} = PM^{-1}$. This allows relating the derivatives: $ \frac{\partial\Phi}{\partial p}_{n\ell} = \sum_n M^{-1}_{\ell n}\frac{\partial \Phi}{\partial \dot{x}}_{kn}$. In matrix form: $D_p\Phi = (I_{3\times 3} \otimes M^{-1}) D_{\dot{x}} \Phi$.

\textbf{Joint constraints}

For a joint constraint $\Phi(X_A,X_B) = X_A\tilde{c}^A - X_B\tilde{c}^B$ connecting bodies $A$ and $B$, there is a similar constraint on the velocities. The Jacobian matrix $D_{x,\dot{x}}\Phi \in \mathbb{R}^{2nd(d+1) \times 2d}$ takes the form:
\begin{equation*}
    (D\Phi)_{(k \alpha n) (i)} = \begin{bmatrix} \frac{\partial \phi_{ij}}{\partial x_{k\alpha}^n} & \frac{\partial \dot{\phi}_{ij}}{\partial x_{k\alpha}^n} \\\frac{\partial \phi_{ij}}{\partial \dot{x}_{k\alpha}^n} & \frac{\partial \dot{\phi}_{ij}}{\partial \dot{x}_{k\alpha}^n} \\\end{bmatrix} = \begin{bmatrix} \tilde{c}_\alpha^A \delta_{ki} \delta_{nA} - \tilde{c}_\alpha^B \delta_{ki} \delta_{nB} &0 \\ 0&\tilde{c}_\alpha^A \delta_{ki} \delta_{nA} - \tilde{c}_\alpha^B \delta_{ki} \delta_{nB} \end{bmatrix}
\end{equation*}
for $k,j=1,..,d$ labeling dimensions, $\alpha=0,...,d$ labeling the extended coordinates, and $n =a,b,...$ labeling the extended bodies. Similarly for the rotational axis restriction.

\section{Benchmark systems}\label{sec:systems}\label{sec:datasets}
Below we detail a series of more challenging synthetic physical systems to benchmark our approach and the baselines. While the information of the Hamiltonian is withheld from our model, we detail the true Hamiltonians below that can be used to generate the data. Even though the equations of motion are very complex, the Hamiltonians in Cartesian coordinates are very simple, again demonstrating why this approach simplifies the learning problem.

\subsection{$N$-Pendulum}
N point masses are connected in a chain with distance constraints $(\vvv{0},\vvv{x}_1), (\vvv{x}_1,\vvv{x}_2), ... ,(\vvv{x}_{N-1}, \vvv{x}_{N})$ and the Hamiltonian is just the contributions from the kinetic energy and gravity in 2 dimensions:
\begin{equation}
    H = \sum_n\vvv{p}_{n}^\top \vvv{p}_{n}/2m_n+gm_nx_{n,2}.
\end{equation}

\subsection{$N$-Coupled pendulums}
In this 3 dimensional system, N point masses are suspended in parallel and springs connect the neighbors horizontally. The distance constraints are $(\vvv{v},\vvv{x}_1), (2\vvv{v},\vvv{x}_2), (3\vvv{v},\vvv{x}_3), ..., (N\vvv{v},\vvv{x}_N)$ where $\vvv{v} = [1,0,0]^\top$ is a horizontal translation (of the origin).
The Hamiltonian is:
\begin{equation}
    H = \sum_{n=1}^N (\vvv{p}_{n}^\top \vvv{p}_{n}/2m_n+gm_nx_{n,2}) + \sum_{n=1}^{N-1} \frac{1}{2}k (\|\vvv{x}_n-\vvv{x}_{n+1}\|-\ell_0)^2
\end{equation}
where $\ell_0=\|\vvv{v}\| = 1$.

\subsection{Magnetic pendulum}
A magnet is suspended on a pendulum (in 3-dimensions) over a collection of magnets on the ground. The pendulum chaotically bounces between the magnets before finally settling on one of them.

Each of the magnets are modeled as dipoles with moments $\vvv{m}_i \in \mathbb{R}^3$. The Hamiltonian for the system is 
\begin{equation}
    H(x,p) = \vvv{p}^\top \vvv{p}/2m - \vvv{m}_0(x)^\top \vvv{B}(x)
\end{equation}
where %
\begin{equation}
    \vvv{B}(x) = \sum_i L(\vvv{x}-\vvv{r}_i)\vvv{m}_i \ \mathrm{and} \ L(r) = \frac{\mu_0}{4\pi \|\vvv{r}\|^5}(3\vvv{r}\vvv{r}^\top -\|\vvv{r}\|^2I), %
\end{equation}
$\vvv{m}_0(x) = -q\frac{\vvv{x}}{\|\vvv{x}\|}$, $\vvv{m}_i = q\hat{\vvv{z}}$ for some magnet strength $q$. $\vvv{r}_i$ are the spatial arrangements of the magnets placed on the plane. %
The constraints are just the distance constraint $(\vvv{0},\vvv{x})$.

\subsection{Gyroscope}\label{sec:gyroscope}
Consider a spinning top that contacts the ground at a single point. To simplify the learning problem, we choose the control points $x_i$ as unit vectors from the center of mass along the principle axes of the top. The Hamiltonian for the system is $H(x,p) = T+V = \Tr(PM^{-1}P^\top )/2 + mg X_{30}$.
where
\begin{align}
    M^{-1}=\frac{1}{m}
    \begin{bmatrix} 1&1&1&1\\
    1& 1+1/\lambda_1&1&1\\1&1&1+1/\lambda_2&1\\
    1&1&1&1+1/\lambda_3
    \end{bmatrix}.
\end{align}
Here we calculate the ground truth moments from the object mesh shown in \autoref{subfig:gyroscope}. In addition to the rigid body constraints, we simply need to add a universal joint connected to the origin.
\subsection{Rigid rotor}
The Hamiltonian of this system in Cartesian coordinates is just the kinetic energy: $H(X,P) = \Tr(PM^{-1}P^\top )/2$. Like for the Gyroscope, we compute the ground truth moments from the object mesh shown in \autoref{subfig:rotor}.
\section{Simplicity of Cartesian coordinates}\label{sec:generalized_derivation}

\subsection{Gyroscope}
The Hamiltonian of a gyroscope in Euler angles $(\phi,\theta,\psi)$ is given by
\begin{align}
\mH = \frac{1}{2} \p^TM(\phi,\theta,\psi)^{-1}p +mg\ell \cos \theta
\end{align}
where the matrix $M$ can be derived by expanding out $\sum_i\mathcal{I}_i\omega_i^2$ using the angular velocity
\begin{equation*}
    \mathbf{\omega} = [\dot{\phi} \sin \theta \sin \psi + \dot{\theta} \cos \psi, \dot{\phi}\sin\theta\cos\psi-\dot{\theta}\sin\psi,\dot{\phi}\cos \theta + \dot{\psi}]^\top
\end{equation*}
yielding the matrix
\begin{align}
M = \begin{bmatrix} \sin^2 \theta (\mathcal{I}_1 \sin^2\psi + \mathcal{I}_2 \cos^2 \psi) +\cos^2\theta I_3 & (\mathcal{I}_1-\mathcal{I}_2)\sin \theta \sin \psi \cos \psi & \mathcal{I}_3 \cos \theta \\ (\mathcal{I}_1-\mathcal{I}_2)\sin \theta \sin \psi \cos \psi & \mathcal{I}_1 \cos^2\psi + \mathcal{I}_2\sin^2\psi & 0\\
\mathcal{I}_3\cos \theta & 0 & \mathcal{I}_3 \end{bmatrix}.
\end{align}
Meanwhile, the Hamiltonian in Cartesian coordinates is given by the simpler form $H(x,p) = T+V = \Tr(PM^{-1}P^\top )/2 + mg X_{30}$, where
\begin{align}
    M^{-1}=\frac{1}{m}
    \begin{bmatrix} 1&1&1&1\\
    1& 1+1/\lambda_1&1&1\\1&1&1+1/\lambda_2&1\\
    1&1&1&1+1/\lambda_3
    \end{bmatrix}.
\end{align}

\subsection{$N$-Pendulum} \label{sec:pendulum}
Suppose we have $N$ linked pendulums in two dimensions indexed from top to bottom with the top pendulum as pendulum $j$. Each pendulum $j$ has mass $m_j$ and is connected to pendulum $j-1$ by a rigid rod of length $l_j$. Let positive $y$ correspond to up and positive $x$ correspond to right. Then
\begin{align*}
    x_j &= \sum_{k=1}^j l_k \sin \theta_k  = \sum_{k=1}^N \mathbbm{1}_{[k\leq j]} l_k \sin \theta_k\\
    y_j &= -\sum_{k=1}^j l_k \cos \theta_k = -\sum_{k=1}^N \mathbbm{1}_{[k\leq j]} l_k \cos \theta_k\\
    \dot x_j &= \sum_{k=1}^j l_k \dot \theta _ k \cos \theta_k = \sum_{k=1}^N \mathbbm{1}_{[k\leq j]}l_k \dot \theta _ k \cos \theta_k\\
    \dot y_j &= \sum_{k=1}^j l_k \dot \theta _ k \sin \theta_k = \sum_{k=1}^N \mathbbm{1}_{[k\leq j]}l_k \dot \theta _ k \sin \theta_k\\
    \frac{\partial \dot x_j}{\partial \dot \theta_i} &= \mathbbm{1}_{[i\leq j]}l_i\cos\theta_i \\
    \frac{\partial \dot y_j}{\partial \dot \theta_i} &= \mathbbm{1}_{[i\leq j]}l_i\sin\theta_i\\ %
\end{align*}
Then given that $T = \sum_{j=1}^N \frac{1}{2}m_j (\dot x_i^2 + \dot y_i^2)$, we have that
\begin{align*}
    p_i &= \frac{\partial T}{\partial \dot \theta_i}\\
        &= \sum_{j=1}^N  m_j\left(\dot x_j \frac{\partial \dot x_j}{\partial \dot \theta_i} + \dot y_j \frac{\partial \dot y_j}{\partial \dot \theta_i}\right) \\
        &= \sum_{j=1}^N m_j \left((\sum_{k=1}^N \mathbbm{1}_{[k\leq j]}l_k \dot\theta_k \cos \theta_k)(\mathbbm{1}_{[i \leq j]}l_i \cos\theta_i) + (\sum_{k=1}^N \mathbbm{1}_{[k\leq j]}l_k \dot\theta_k \sin \theta_k)(\mathbbm{1}_{[i \leq j]}l_i \sin\theta_i)\right) \\
        &= \sum_{j=1}^N m_j\mathbbm{1}_{[i \leq j]}\left(\sum_{k=1}^N \mathbbm{1}_{[k\leq j]}l_kl_i (\cos \theta_k \cos\theta_i + \sin \theta_k \sin\theta_i)\dot \theta_k\right) \\
        &= \sum_{j=1}^N \sum_{k=1}^N m_j \mathbbm{1}_{[k\leq j]}\mathbbm{1}_{[i \leq j]} l_i l_k \cos(\theta_i - \theta_k) \dot\theta_k\\
        &= \sum_{k=1}^N \sum_{j=1}^N m_j \mathbbm{1}_{[k\leq j]}\mathbbm{1}_{[i \leq j]} l_i l_k \cos(\theta_i - \theta_k) \dot\theta_k\\
        &=\sum_{k=1}^N  \left(l_i l_k \cos(\theta_i - \theta_k)\sum_{j=1}^N m_j\mathbbm{1}_{[k \leq j]}\mathbbm{1}_{[i \leq j]}\right)\dot\theta_k  \\
        &= \sum_{k=1}^N  \left(l_i l_k \cos(\theta_i - \theta_k)\sum_{j=\max(i, k)}^N m_j\right)\dot\theta_k 
\end{align*}
where we made use of the fact that $\cos\theta_k\cos\theta_i + \sin\theta_k \sin\theta_i = \cos(\theta_i - \theta_k)$.
To obtain the entries of the mass matrix, we match the above equation to for $p_i$ with the expected form
\begin{align*}
    p_i &= \sum_{k=1}^N M_{ik} \dot\theta_k
\end{align*}
which gives
\begin{align*}
    M_{ik} &= l_i l_k \cos(\theta_i - \theta_k)\sum_{j=\max(i,k)}^N m_j.
\end{align*}

\textbf{Equations of motion for the $N$-Pendulum.}
For the $N=2$ case, the equations of motion in generalized coordinates are
\begin{align*}
     \dot q_1 &= \frac{\ell_2 p_1 - \ell_1 p_2 \cos (q_1 - q_2)}{\ell_1^2 \ell_2 (m_1 + m_2 \sin^2(q_1-q_2))}
     &   \dot q_2 &= \frac{-m_2\ell_2 p_1 \cos(q_1 - q_2) + (m_1 + m_2) \ell_1 p_2}{m_2\ell_1\ell_2^2(m_1 + m_2 \sin^2 (q_1-q_2))} \\
     \dot p_1 &= -(m_1+m_2)g\ell_1\sin \theta_1 - C_1 + C_2
     &   \dot p_2 &= -m_2g\ell_2 \sin q_2 + C_1 - C_2 
\end{align*}
where
\begin{align*}
    C_1 &= \frac{p_1p_2\sin(q_1-q_2)}{\ell_1\ell_2(m_1+m_2\sin^2(q_1-q_2))} & C_2 = \frac{m_2\ell_2^2p_1^2 + (m_1+m_2)\ell_1^2p_2^2 - 2m_2\ell_1\ell_2p_1p_2\cos (q_1-q_2)}{2\ell_1^2\ell_2^2(m_1+m_2\sin^2(q_1-q_2))^2/\sin(2(q_1-q_2))}.
\end{align*}
For $N=3$, \textit{the equations of motion would stretch over a full page}.

On the other hand the equations of motion in Cartesian coordinates are described simply by
\begin{align*}
    \quad \dot x_{i,1} = \frac{p_{i,1}}{m_i}
    \quad \dot x_{i,2} = \frac{p_{i,2}}{m_i}
    \quad \dot p_{i,1} = 0
    \quad \dot p_{i,2} = gm_i
\end{align*}
which maintain the same functional form irrespective of $N$.

\end{document}